\documentclass[]{bytedance_seed}
\usepackage[toc,page,header]{appendix}
\usepackage{minitoc}
\usepackage{textpos}
\usepackage{amssymb}
\crefformat{figure}{Fig. #2#1#3}
\crefformat{table}{Table #2#1#3}

\title{Learning Vision-Driven Reactive Soccer Skills for Humanoid Robots}

\author[1,2,*]{Yushi Wang}
\author[1]{Changsheng Luo}
\author[1,2,*]{Penghui Chen}
\author[2]{Jianran Liu}
\author[2]{Weijian Sun}
\author[1]{Tong Guo}
\author[3]{Kechang Yang}
\author[3]{Biao Hu}
\author[2]{Yangang Zhang}
\author[1,\dagger]{Mingguo Zhao}

\affiliation[1]{Tsinghua University}
\affiliation[2]{ByteDance Seed}
\affiliation[3]{China Agricultural University}

\contribution[*]{Work done at ByteDance Seed}
\contribution[\dagger]{Corresponding author}

\abstract{
Humanoid soccer poses a representative challenge for embodied intelligence, requiring robots to coordinate agile locomotion with unreliable visual perception in dynamic environments. However, existing systems typically rely on modular pipelines that separate perception from control or assume ideal sensing, making it difficult to achieve coherent and reactive behavior under real-world perceptual limitations. In this work, we present a unified reinforcement learning-based controller that enables humanoid robots to learn vision-driven reactive soccer skills by directly coupling visual perception with locomotion control. The robot is trained in simulation to acquire soccer behaviors, and adversarial motion priors guide policy learning toward natural motion patterns. To support robust performance under imperfect sensing, we introduce an encoder-decoder architecture together with a virtual perception system that models key characteristics of onboard vision, exposing the policy to perceptual noise and detection failures during training. This design encourages the policy to internalize perceptual uncertainty and continuously adapt its motion in a closed loop. The resulting controller produces coordinated soccer behaviors using only onboard vision, including ball searching, chasing, and multidirectional kicking. It reduces ball position estimation error by 46\% and shortens time-to-kick by up to 64\% compared with a rule-based baseline, achieving around 90\% kicking success in frontfield positions. Experiments across diverse environments and dynamic scenarios, including real RoboCup competitions, further demonstrate the robust performance of the controller. These results highlight the practical effectiveness of integrating perceptual uncertainty directly into policy learning for achieving reliable vision-driven behaviors in humanoid robots operating under real-world conditions.
}

\date{\today}
\correspondence{Mingguo Zhao at \email{mgzhao@tsinghua.edu.cn}}
\checkdata[Project Page]{\url{https://humanoid-kick.github.io}}

\begin{document}
\maketitle

\section{Introduction}
Soccer stands as a globally celebrated sport that epitomizes the pinnacle of human sensorimotor intelligence. Mastery of the game demands not only physical prowess but also an intuitive perception-action coupling that enables a seamless fusion of acute visual perception, rapid decision-making, and adaptive sensorimotor coordination. Human players track a fast-moving ball, anticipate opponents' actions, adjust their posture midstride, and strike with precise timing and force. These capabilities are precisely what humanoid robots require to operate autonomously in unstructured, dynamic environments. Unlike benchmarks that target either locomotion or manipulation skills, robot soccer serves as a focused testbed that compresses a broad spectrum of essential robotic competencies into a single continuous task, offering a practical arena that drives technological advances toward human-level embodied intelligence in the real world.

However, reproducing even the most basic soccer skills on humanoid robots remains a substantial challenge. The robot is required to generate long-horizon behaviors that satisfy complex task demands while continuously adapting its motion in real time to rapidly changing sensory feedback. These challenges are further amplified by onboard visual perception that is noisy, delayed, and spatially constrained during dynamic motion, leading to distorted or incomplete observations of the environment state. The controller must remain robust and generalizable when exposed to disturbances introduced by competitive physical interactions and visually cluttered environments. To overcome these obstacles, the robot requires advanced locomotion capabilities, reliable visual tracking of dynamic targets, and rapid adjustment of foot placement and body orientation, ultimately achieving successful strikes through precise synchronization of timing, angle, and force. Achieving such tightly coupled perception-action competence goes far beyond conventional locomotion control and becomes even more challenging under real-world deployment constraints.

Robot soccer has attracted substantial research attention, particularly under the auspices of RoboCup \cite{kitano1997robocup}. Classical systems typically adopt modular pipelines that decompose the problem into separate perception, planning, and control components \cite{fernandez2024robocup}. Although such designs offer engineering convenience and interpretability, they often require extensive manual tuning and struggle to achieve agile and reactive behaviors in dynamic environments, given that effective responses rely heavily on accurate perception estimates. Learning-based approaches have also been explored for robot soccer, with some studies focusing on isolated skills such as running \cite{abreu2019learning}, dribbling \cite{bohez2022imitate,ji2023dribblebot,ravichandar2024preferenced,shao2025visualmimic}, shooting \cite{ji2022hierarchical}, or goalkeeping \cite{huang2023creating}, whereas others target high-level strategic planning such as team coordination \cite{abreu2025designing,su2025toward}. However, these methods largely continue to decompose soccer into separate skills, limiting their ability to produce integrated and reactive behaviors.

A major paradigm shift emerged with the introduction of end-to-end reinforcement learning (RL) for humanoid soccer, where unified policies were trained to acquire long-horizon behaviors \cite{haarnoja2024learning}. This study demonstrated the feasibility of learning compound soccer behaviors through RL and represented a milestone for humanoid sports research. However, the learned policy relied on privileged state information provided by a motion capture system, which precluded deployment beyond laboratory environments and limited its applicability in real-world settings. Although the subsequent work \cite{tirumala2024learning} extended this framework toward vision-based control by incorporating egocentric perception rendered through neural radiance fields \cite{mildenhall2021nerf} during training, it still depended on computationally intensive scene-specific modeling and training, restricting scalability to real-world scenarios. Moreover, the sim-to-real transfer exhibited noticeable degradation in reactivity and success rate, highlighting the persistent challenges of robust visual perception and control coupling in physical humanoid systems. Consequently, a vision-driven reactive humanoid controller capable of robust soccer behaviors in real-world environments remains an open challenge.

Beyond soccer, RL has driven advances in humanoid control, enabling robust and agile behaviors such as running \cite{crowley2025optimizing}, jumping \cite{li2025reinforcement}, parkour \cite{zhuang2024humanoid}, and fall recovery \cite{huang2025learning,chen2025hifar}. Recently, learning from demonstrations has emerged as a popular framework, empowering humanoid robots to replicate specific human behaviors in an expressive, humanlike manner, including stylized walking, dancing, leg kicks, and acrobatic flips \cite{zhang2025hub,xie2025kungfubot,chen2025gmt,yin2025unitracker,zeng2025behavior,zhang2025track}. By providing structured guidance from expert data, these approaches substantially reduce exploration complexity and bias policy optimization toward desired behaviors \cite{li2025feature}. The most widely adopted feature-based approaches, established by DeepMimic \cite{peng2018deepmimic}, construct dense tracking rewards on the basis of handcrafted motion features to achieve high-fidelity imitation. Although effective in reproducing detailed motion patterns, such formulations rely on rigid temporal alignment and exhibit reduced adaptability. Subsequent work sought to improve flexibility via supervised distillation \cite{allshire2025visual}, hierarchical generation of reference motions \cite{truong2025beyondmimic}, or latent motion embeddings \cite{bohez2022imitate,zhang2025motion,zhang2025unleashing,kang2025learning}, often at the cost of increased system complexity and multistage training pipelines. In contrast, generative adversarial network (GAN)-based approaches \cite{peng2021amp, escontrela2022adversarial, vollenweider2022advanced,alvarez2025learning,wen2025constrained} have introduced another alternative by learning implicit motion distributions from expert demonstrations and integrating imitation with task optimization. These methods avoid explicit temporal alignment and allow flexible adaptation to task objectives, with further improvements achieved using Wasserstein GAN formulations \cite{arjovsky2017wasserstein,gulrajani2017improved,tang2024humanmimic} to mitigate limitations like discriminator saturation and mode collapse. Despite their promise, existing studies primarily focus on motion imitation using proprioceptive or privileged state information, overlooking the integration of exteroceptive sensing and perception-action coupling, which are indispensable for dynamic and interactive tasks such as robot soccer.

In this work, we formulate humanoid soccer as a perception-constrained control problem. Moving beyond modular pipelines and idealized perception assumptions, we explicitly model realistic perceptual uncertainty as an integral component of the learning dynamics rather than treating perception as a preprocessing module that provides inputs to the controller. To this end, we developed a vision-driven reactive control framework that tightly couples visual perception with locomotion, enabling the policy to internalize uncertainty and continuously adapt its motion in a closed loop. The robot is trained via RL to kick a ball toward the goal in a simulated soccer environment, and adversarial motion priors (AMP) \cite{peng2021amp} guide policy optimization toward expert-like motion patterns. We designed a compact perception abstraction that extracts task-relevant structured cues from raw visual inputs and fuses them with odometry to support long-horizon spatial awareness. Built on this abstraction, a virtual perception system emulates key characteristics of onboard vision, thereby exposing the policy to realistic sensing imperfections during simulation training. An encoder-decoder architecture further compresses noisy observations into compact latent representations for control, forcing the policy to internalize imperfect sensory feedback and implicitly infer privileged states in its latent dynamics. This design allows perceptual uncertainty to directly influence locomotion strategy and motion generation, resulting in adaptive and reactive behaviors under real-world sensing constraints. Collectively, our work demonstrates that a unified policy can effectively couple perception and locomotion to achieve robust vision-driven soccer behaviors on physical humanoid robots.

Extensive experiments validated the controller's ability to execute agile and reactive soccer behaviors across diverse scenarios using only onboard vision, including ball searching, chasing, and multidirectional kicking. The policy actively searched for and tracked the ball while implicitly compensating for perception noise, exhibiting tight perception-action coordination. By dynamically adapting its gait to the ball's position, the policy achieved fast and powerful shots while maintaining stability. As the champion team of the Adult-Size Humanoid League in RoboCup 2025 and the 2025 World Humanoid Robot Games, we further showcased the practical efficacy of the proposed method in real-world competitive settings, where the robot operated under strict sensing and interaction constraints yet delivered a strong performance.

\section{Results}

We evaluated the proposed learning framework through extensive simulations and real-world experiments. The results demonstrated that the learned policy enabled reactive soccer skills, seamlessly integrating real-time ball tracking, agile locomotion, and accurate kicking while maintaining robust performance across diverse scenarios. We further analyzed emergent perceptual and gait behaviors to reveal how the policy coupled visual cues with motion control.

\subsection{System overview}

\begin{figure}[!t]
    \centering
    \includegraphics[width=\textwidth]{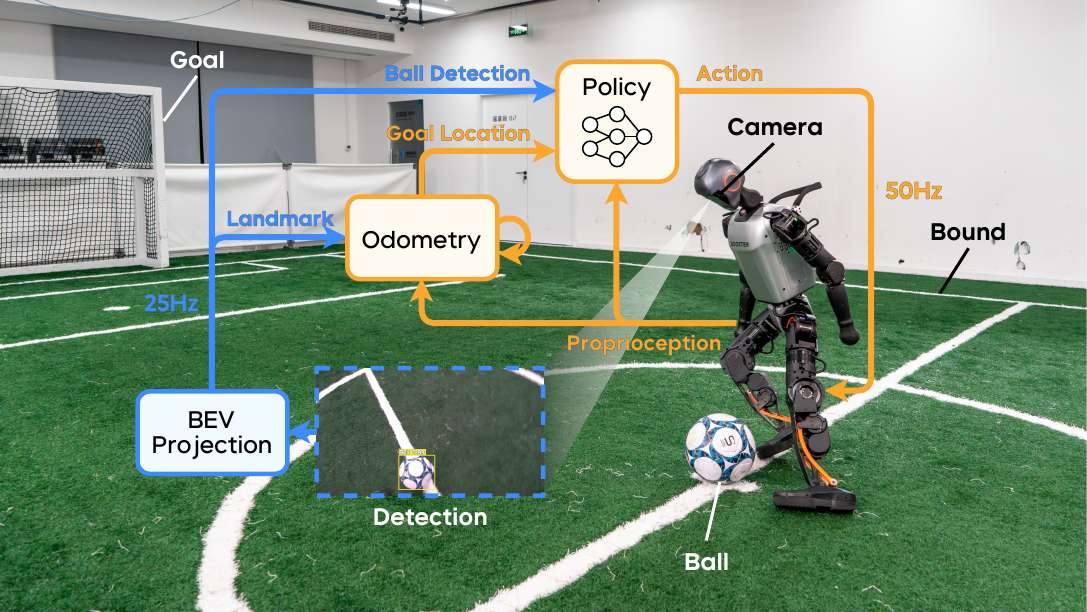}
    \caption{\textbf{System overview.} The real-world robot is equipped with an onboard camera for visual perception. Image detections are projected into the BEV space. Ball detections are provided directly to the policy, and field landmarks are processed by an odometry module to infer the goal location from long-term information. The perception pipeline is designed to efficiently extract and represent visual features for the RL policy.}
    \label{fig:overview}
\end{figure}

We deployed our policy on the Booster T1, a humanoid robot measuring $\sim1.2$~m in height and 30~kg in mass. Booster T1 has been used as a humanoid platform for locomotion control \cite{chen2025hifar}, locomanipulation \cite{zhang2025falcon,jang2025seec}, navigation \cite{cheng2024navila}, and the development of RL algorithms \cite{seo2025fasttd3} and suites \cite{wang2025booster,zakka2025mujoco,geng2025roboverse}. For the present work, no specific modifications were made to the robot's original hardware. The robot is equipped with an Intel RealSense Depth Camera D455i, mounted on its head with two joints to control its yaw and pitch orientation during dynamic movements, allowing for active ball and field perception. We trained the soccer policy in simulation and deployed it directly onto the real robot. As illustrated in Fig.~\ref{fig:overview}, images captured by the camera are processed by the detection pipeline and projected into bird's-eye-view (BEV) coordinates on the onboard Jetson AGX Orin at 25~Hz. The policy receives proprioception data, ball detections, and goal information from a separate odometry module, generating joint position commands at a frequency of 50~Hz.

\subsection{Soccer skills performance}

\begin{figure}[!t]
    \centering
    \includegraphics[width=\textwidth]{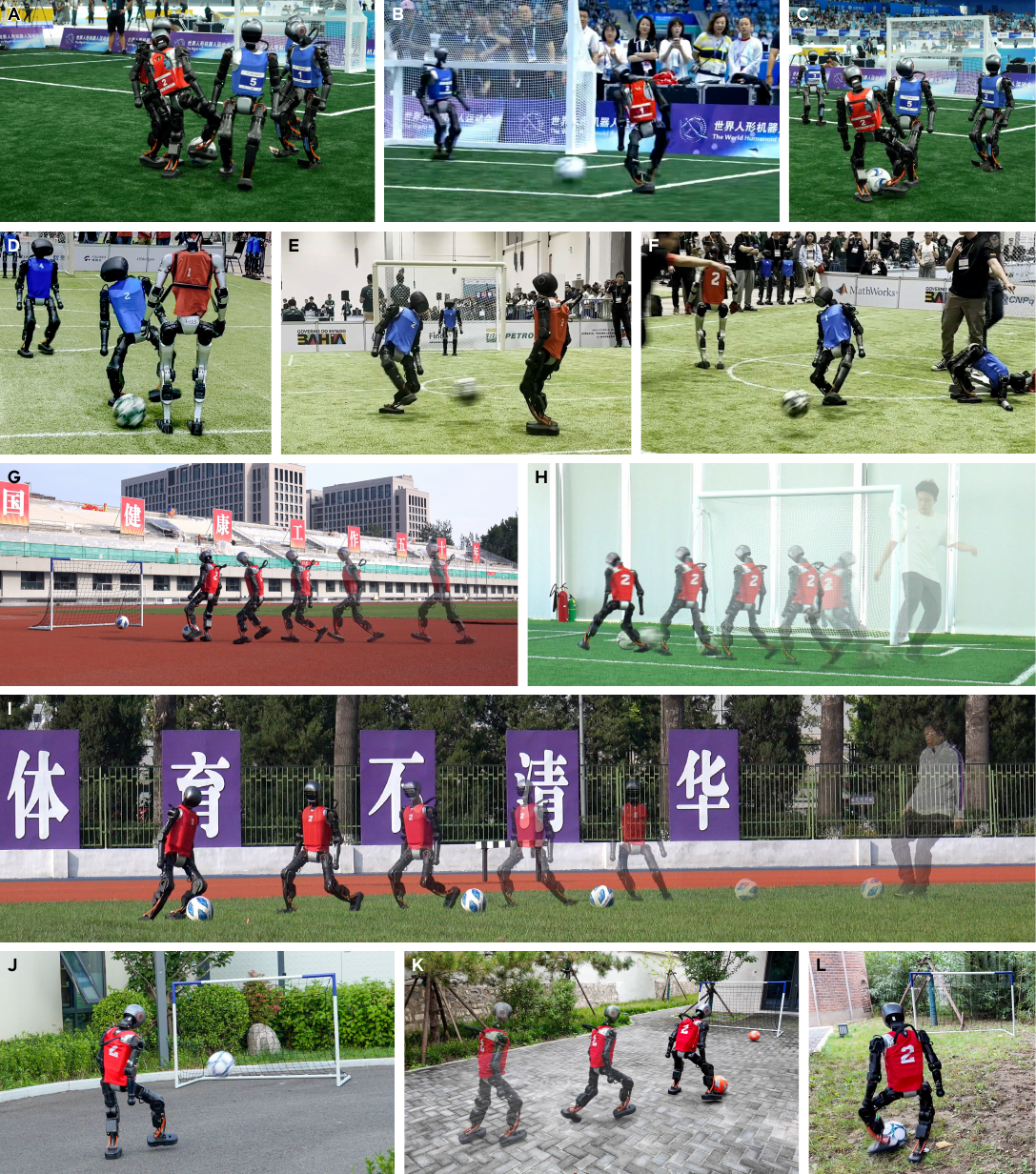}
    \caption{\textbf{Performance of the controller in various scenarios.} (\textbf{A} to \textbf{C}) Real match performance at the World Humanoid Robot Games. (\textbf{D} to \textbf{F}) Real match performance at RoboCup. (\textbf{G} to \textbf{I}) Reactive responses and real-time adaptation to the ball. (\textbf{J} to \textbf{L}) Robust behavior across varying terrain and visually diverse environments.}
    \label{fig:outdoor}
\end{figure}

We deployed the proposed controller across a diverse range of scenarios to assess its ability to execute robust reactive soccer skills, as shown in Fig.~\ref{fig:outdoor}. The evaluation encompassed various surface types, including grass, slabstone, soil, asphalt, and rubber, each characterized by distinct terrain properties that led to varied interaction dynamics. In addition, visual conditions such as color and illumination varied substantially across scenarios, occasionally causing detection failures. Notably, the controller was trained entirely in simulation without incorporating data from these environments and was deployed to the real-world zero-shot. Despite these challenges, the controller demonstrated robust adaptation, consistently achieving reliable ball tracking and locomotion. Even in scenarios lacking field landmarks, the robot sustained robust soccer performance using proprioceptive odometry.

We further validated the system's ability to sustain extended rounds. The policy successfully adapted to multiple balls placed at various locations on the field, enabling continuous kicking with diverse and coherent behaviors. In interactive matches against a human goalkeeper, the robot adjusted its movements to maintain consistent contact with the ball, continuously tracking and executing shots, and even occasionally overcoming the human defense. When ball tracking was temporarily lost, the robot exhibited active visual search behaviors, thereby ensuring continuity in kicking and enabling long-duration interactions.

The proposed controller was adopted as a module by the Tsinghua Hephaestus team in both the RoboCup 2025 Adult-Size Humanoid League and the 2025 World Humanoid Robot Games, where the team won championships with 76 goals scored and only 11 conceded. In these competitions, robots were required to operate fully autonomously, relying exclusively on onboard computation and sensing, and the use of sensors exceeding human sensory capabilities, such as LIDAR (light detection and ranging) or multiple camera systems, was strictly prohibited. Moreover, adaptation time to the actual competition venue was highly limited. These factors collectively presented substantial challenges to the deployment of our method. Despite these constraints, our approach demonstrated its effectiveness and agility, contributing to multiple successful goals during the competitions. By enabling agile kicking motions, this approach allowed the robot to gain an advantage in ball possession against opponents. The method also exhibited robust performance in challenging scenarios, including accurate long-range shots from the touchline midpoint and reliable operation despite visual occlusion and external physical disturbances. To clarify how the proposed controller was integrated into the full RoboCup system, we describe details of the deployment architecture used in real matches in the Appendix.

\subsection{Kicking success rates}

\begin{figure}[!t]
    \centering
    \includegraphics[width=\textwidth]{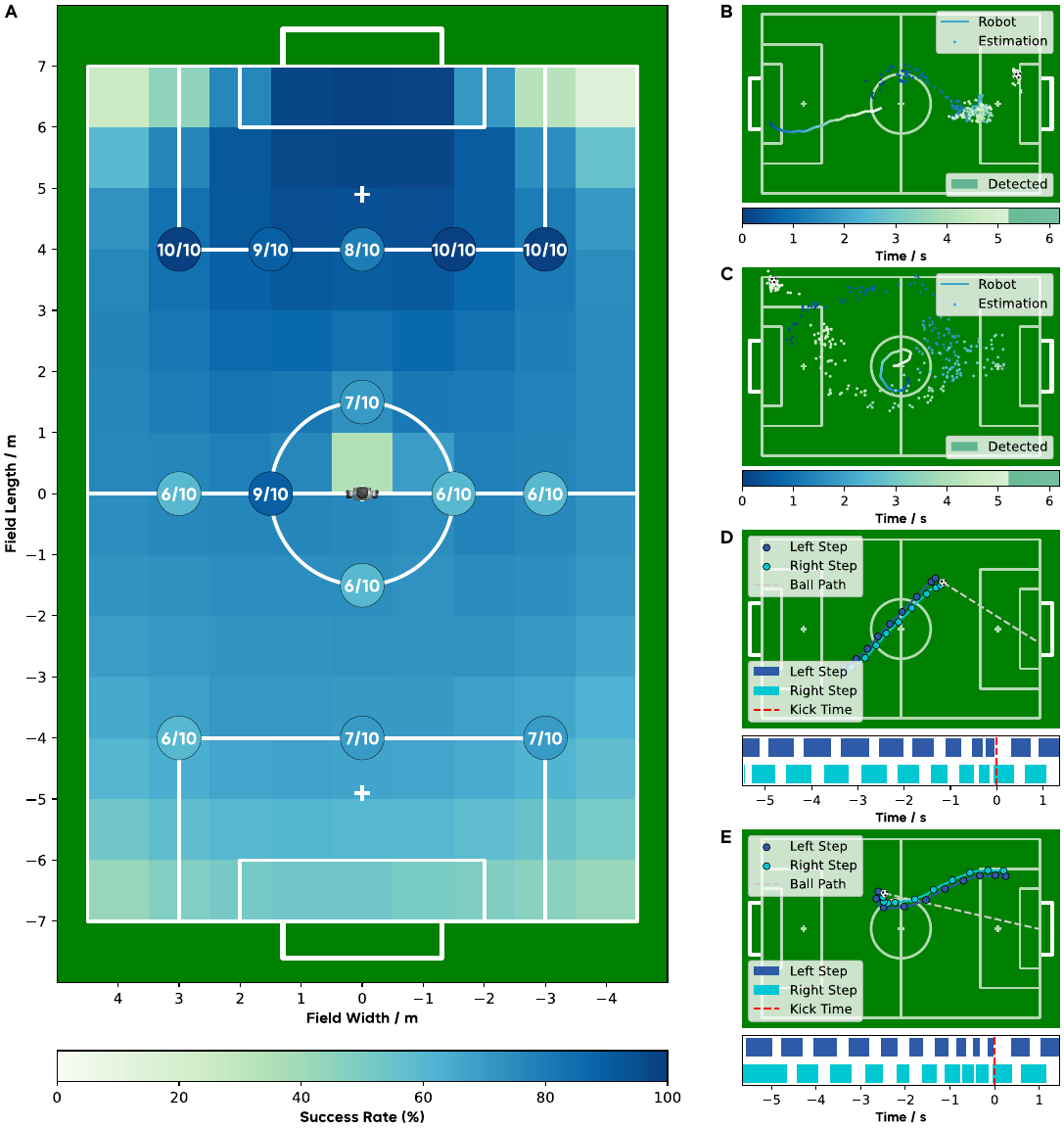}
    \caption{\textbf{Validation and behavior analysis.} (\textbf{A}) The background grid color represents the success rate of 8192 simulation tests, and the circles indicate the success rate of 10 consecutive hardware tests. Owing to effective alignment, our policy delivers reliable hardware performance that closely matches simulation results. (\textbf{B}) The robot searches for the ball in the distance when starting near the field edge, guided by the policy's estimated ball position. (\textbf{C}) The robot turns to search for the ball behind itself when it is near the field center. (\textbf{D} and \textbf{E}) The robot's foothold locations and timing reveal adaptive gaits with shorter strides and faster cadence, enabling effective adjustment before kicking, as illustrated by the forward and backward kick examples.}
    \label{fig:success_rate}
\end{figure}

We evaluated the proposed policy by measuring success rates across different regions of the soccer field, following the adult-size specifications of the RoboCup Humanoid League. At the start of each trial, the ball was placed at predefined positions on the field, whereas the robot was always initialized at the center. A trial was considered successful if the robot kicked the ball into the goal and failed if the robot fell or the ball went out of bounds. If the ball remained in play, the robot was permitted to attempt additional kicks. Consecutive trials were conducted for each region, and the resulting success rates are summarized in Fig.~\ref{fig:success_rate}A.

We first evaluated the policy in simulation. To ensure consistency, external disturbances such as uneven terrain and physical perturbations to the ball or robot were removed, whereas the virtual perception system modeled after real-world visual characteristics was retained. This setup allowed for a comprehensive evaluation of the vision-driven pipeline, from perception to motor execution, requiring the robot to handle perceptual noise, accurately localize the ball, and generate appropriate kicking motions. Results demonstrated the high reliability of the policy in executing soccer skills. In regions close to the goal, success rates were particularly high, reaching $\sim90\%$, with only occasional failures. Performance declined as the distance from the goal or angular offsets relative to the goal's normal direction increased. This reduction was primarily due to tighter tolerance for kicking angle accuracy, especially in backfield regions, where angular deviations exceeding $\pm10^\circ$ resulted in missed shots. In addition, success rates were lower when the ball was positioned directly along the line between the robot and the goal, given that the robot struggled to select the appropriate kicking foot, and when the ball was placed very close to the robot, because the limited time constrained accurate ball localization and gait adaptation. Nevertheless, the robot achieved successful kicks in most of the tested regions.

We further validated the policy on the physical robot. The hardware experiments achieved success rates comparable to those observed in simulation, with $\sim80$ to 90\% success in the frontfield positions and 60 to 70\% success in the more challenging backfield regions over 10 consecutive tests per position. These findings confirmed that the virtual perception system effectively bridged the gap between simulation and the real world. Notably, the robot remained stable throughout all trials, with no falls recorded, underscoring the robustness of the proposed policy. These results collectively demonstrated the effectiveness of our controller in real-world settings.

\subsection{Perception-action coordination}

\begin{figure}[!t]
    \centering
    \includegraphics[width=\textwidth]{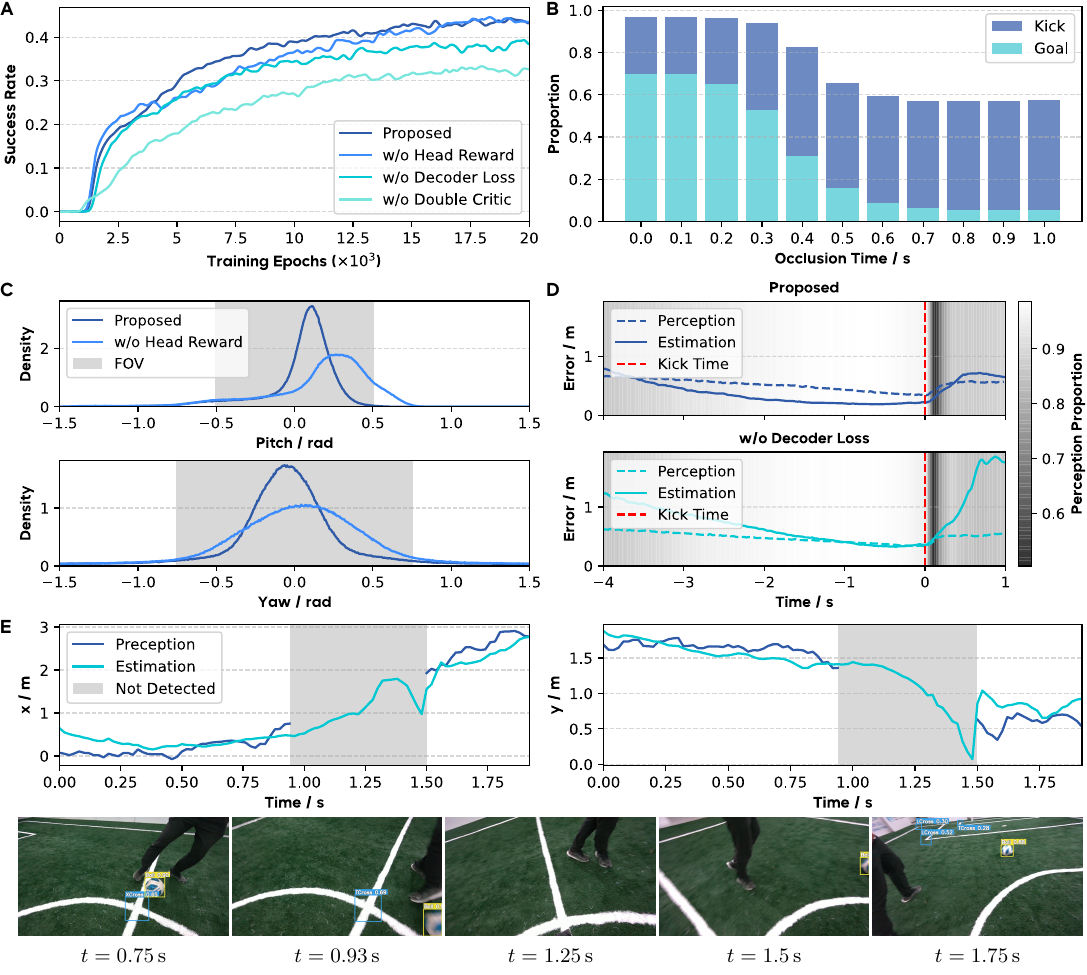}
    \caption{\textbf{Perception-action coordination.} (\textbf{A}) Training curves for different methods, reporting overall success rates in disturbed training environments. (\textbf{B}) Kick and goal-scoring success rates under visual occlusion, evaluated over 8192 simulation trials per occlusion duration. Occlusion is simulated by synthetically canceling ball detections when the robot reaches within 1.4~m of the ball ($\sim1$~s before kicking). (\textbf{C}) Distribution of angular distance between the ball and the camera center, evaluated over 1000 steps across 2048 environments. The shaded area indicates the camera's FOV. (\textbf{D}) Proportion of ball perception, average perception error, and policy's ball position estimation error across 4096 kicking tests. (\textbf{E}) The policy's estimate predicts the ball's movement when the ball is removed from the robot's FOV, guiding the robot to reacquire visual detection of the ball along the direction in which it disappeared.}
    \label{fig:reconstruct_noise}
\end{figure}

We investigated the policy's perception-action coordination, emphasizing its active adjustment of the robot's torso and head orientations to sustain robust ball tracking under partially observable perceptual conditions. As illustrated in Fig.~\ref{fig:reconstruct_noise}C, this behavior enabled the policy to keep the ball in the camera's field of view (FOV) for the majority of the time. This active perceptual behavior, together with successful kick execution, was preserved even when explicit ball-tracking rewards were removed. However, incorporating such rewards encouraged the robot to maintain the ball near the center of its FOV rather than at the periphery, thereby improving robustness in real-world deployment. We further examined the active perception capability in dynamic scenarios. As shown in Fig.~\ref{fig:reconstruct_noise}E, when the ball rolled toward the edge of the FOV, the policy adaptively rotated the robot's head and torso toward the direction in which the ball was leaving the view, successfully repositioning it in the camera's view to sustain tracking. The ball positions estimated by the decoder network in the policy confirmed that the robot deliberately oriented itself toward the predicted trajectory of ball motion.

We further analyzed the policy's ball position estimation and its alignment with the ground-truth position. As shown in Fig.~\ref{fig:reconstruct_noise}D, compared with the raw visual perception, the policy's estimates exhibited substantially reduced noise, with the minimum root mean square error in the last 1~s before kicking decreasing from 0.344 to 0.186~m. This demonstrated the policy's effectiveness in denoising perceptual data. Considering that the length of the robot's arch is only 0.23~m, such an improvement is essential for accurate ball positioning and reliable strikes. This capability arose from the encoder-decoder architecture in the policy network, where latent states retain ball information over a 1-s observation window to achieve accurate estimates. If the decoder was removed during policy training and trained separately afterward to estimate the ball position from latent states, the resulting predictions remained at the noise level.

To further evaluate the robustness of this perception-action coordination under transient perceptual failures, we examined the policy's responsiveness to sudden visual occlusions. Specifically, we artificially disabled ball detection inputs for durations ranging from 0.1 to 1.0~s before kicking in simulation. Given that the policy receives object-centric detection outputs rather than raw images, the synthetic occlusion is equivalent to temporary loss of ball observations caused by real visual occlusions at the policy input level. As shown in Fig.~\ref{fig:reconstruct_noise}B, when the occlusion duration was no more than 0.3~s, the robot successfully kicked the ball in more than 90\% of trials and maintained a goal-scoring success rate exceeding 50\%. For longer occlusions, the robot failed to kick the ball in $\sim40\%$ of trials, whereas the remaining trials still resulted in kicks but with markedly reduced precision. Similar qualitative behaviors were also observed in real-world experiments. Under short visual occlusions of $\sim0.3$~s, the robot continued chasing the ball and successfully completed the kick. In contrast, for longer occlusions of around 0.6~s, the robot lost track of the ball after the occlusion and switched to rotational search behaviors, likely because the prolonged absence of observations led the policy to interpret the ball as having left the FOV. These results suggested that the policy retained short-term memory under temporary loss of visual observations.

During the ball-search phase, the policy also exhibited an adaptive strategy tailored to field position, as illustrated in Fig.~\ref{fig:success_rate} (B and C). When the robot was near the field edges, it first reoriented toward the center, enabling a wide visual sweep of the playing area. It then tended to move toward the central region, which provided a positional advantage for detecting balls located in distant corners or along the perimeter. Upon reaching the center, the robot switched to a rotational scanning pattern, covering all spatial zones to ensure full visual coverage. The searching strategy was strongly supported by the policy's ball position estimates during the search. The predicted coordinates consistently corresponded to field regions to be explored. For example, while moving toward the center, the estimates often pointed to distant outer areas, guiding the robot to expand its search range. Similarly, when turning to scan behind, the estimates shifted to rearward regions not yet surveyed. This close coupling between search behaviors and predictive estimates highlighted the policy's capacity to integrate spatial awareness of unexplored areas into decision-making, thereby improving ball localization efficiency.

\subsection{Gait behavior analysis}

\begin{figure}[!t]
    \centering
    \includegraphics[width=\textwidth]{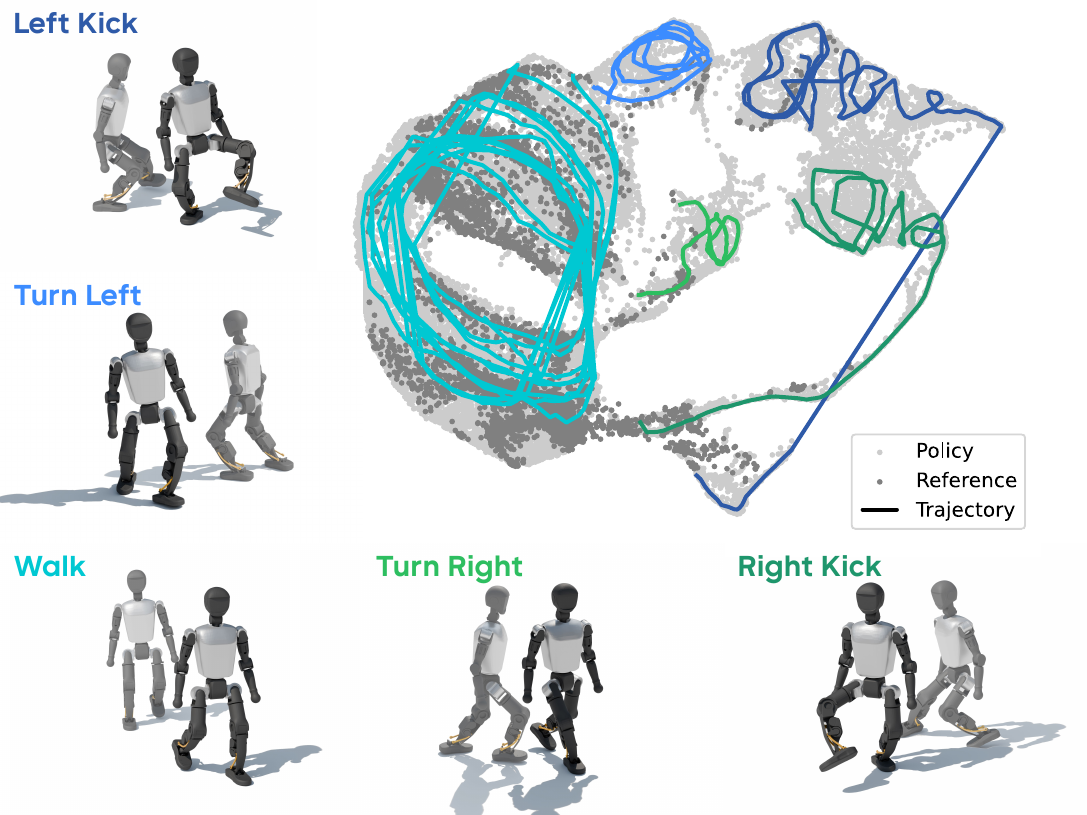}
    \caption{\textbf{Versatile gait behaviors.} Visualization of 20,000 collected joint-space trajectory frames together with the reference motion dataset reduced to a 2D plane using UMAP. Five distinct clusters highlight the policy's ability to integrate reference motions and generalize beyond them to generate task-specific behaviors. From each major cluster, a representative trajectory is selected to illustrate the corresponding behavior.}
    \label{fig:umap}
\end{figure}

We analyzed the gait behaviors exhibited by our controller. Following the approach in \cite{haarnoja2024learning}, we applied Uniform Manifold Approximation and Projection (UMAP) \cite{mcinnes2018umap} to reduce the dimensionality of the joint-space trajectories generated by our policy into a two-dimensional (2D) space for visualization. As shown in Fig.~\ref{fig:umap}, the resulting embeddings revealed five distinct gait clusters, corresponding to walking, turning left, turning right, left-foot kicking, and right-foot kicking. These results demonstrated that our controller produced a versatile set of behaviors in a single policy and transitioned smoothly among them, enabling coherent execution of tasks such as searching, chasing, and kicking the ball. To further evaluate the effectiveness of AMP, we projected the reference motion dataset into the same space. The policy trajectories broadly covered the dataset, indicating successful integration of demonstrated motions, and extended beyond it to synthesize task-specific behaviors. A representative example was the pivot hook kick observed when the goal was behind the robot. In this case, the robot pivoted on the supporting foot and swung the kicking leg laterally to hook the ball, eliminating the need for a full-body rotation and thereby reducing execution time. This underscored the policy's capacity to compose novel behaviors beyond demonstrated motions, effectively adapting to task requirements.

\begin{figure}[!t]
    \centering
    \includegraphics[width=\textwidth]{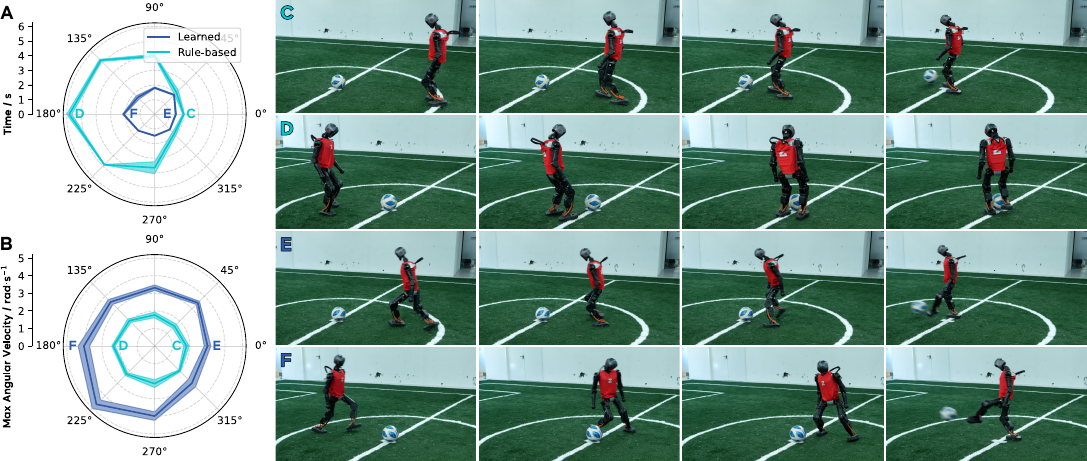}
    \caption{\textbf{Comparison of the learned policy versus a rule-based strategy representative of the current RoboCup state of the art.} (\textbf{A} and \textbf{B}) Time from robot start to ball contact and the maximum angular velocity of the robot during this process, measured across various approach angles. The robot starts stationary from a 1.5-m distance. Shaded regions indicate the SD, calculated from five tests per direction. (\textbf{C} and \textbf{D}) Representative behaviors when the robot kicks the ball forward ($0^\circ$) and backward ($180^\circ$) with the rule-based strategy. (\textbf{E} and \textbf{F}) The learned policy enables the robot to integrate approaching and kicking the ball.}
    \label{fig:kick_time}
\end{figure}

We compared the learned policy with a state-of-the-art rule-based soccer strategy used by the runner-up team in the RoboCup Humanoid League. The rule-based approach leveraged the robot's built-in walking controller and relied on preprogrammed behavior trees to generate velocity commands for decision-making. In our experiment, the ball was positioned at the center of the field, and the robot was initialized on a circle with a 1.5-m radius centered on the ball. As illustrated in Fig.~\ref{fig:kick_time}, we quantitatively evaluated performance in terms of kicking time and agility. Kicking time was defined as the interval between the onset of movement and the moment the ball was kicked out. Using the rule-based strategy, the process required $\sim2$~s when the robot was oriented toward the goal but increased to about 5~s when the target direction was located behind the robot. This delay resulted from the need for extensive rotation and fine positional adjustments around the ball to achieve proper alignment. In contrast, the learned policy consistently achieved shorter times of $\sim1.5$~s across all orientations, with only minor variation. This improvement could be attributed to its capacity for dynamically adjusting foot placement, enabling a seamless transition from approaching the ball to executing the kick. Notably, the policy integrated the kicking motion into the final step of the approach gait rather than preceding it with a distinct preparatory phase. This compact motion pattern minimized execution delay and facilitated reactive kick behavior during ball interactions. We further analyzed the robot's maximum angular velocity during the kicking process to assess agility. Although not a complete measure of agility, this metric reflected the presence of agile behavior. The learned policy attained higher turning speeds of 3 to 4~rad/s across all tested orientations, compared with $\sim2$~rad/s for the rule-based strategy, underscoring its greater agility in aligning with the target direction.

We further analyzed the gait characteristics exhibited by the policy during ball tracking, revealing adaptive patterns tailored to the ball's proximity and movement state. As illustrated in Fig.~\ref{fig:success_rate} (D and E), when the robot was at a greater distance from the ball, it used a gait with a slower step frequency, leading to a gait cycle duration of $\sim0.7$ to 0.8~s, allowing the robot to maintain consistent visual tracking of the ball while conserving energy. In contrast, as the robot approached the ball, a notable increase in step frequency was observed, with the gait cycle duration reduced to $\sim0.3$ to 0.4~s, producing shorter and more rapid strides for accurate adjustment of foot placement in the vicinity of the ball. This accelerated gait pattern enhanced the robot's maneuverability, enabling it to make subtle positional corrections to ensure alignment with the optimal kicking position.

\begin{figure}[!t]
    \centering
    \includegraphics[width=\textwidth]{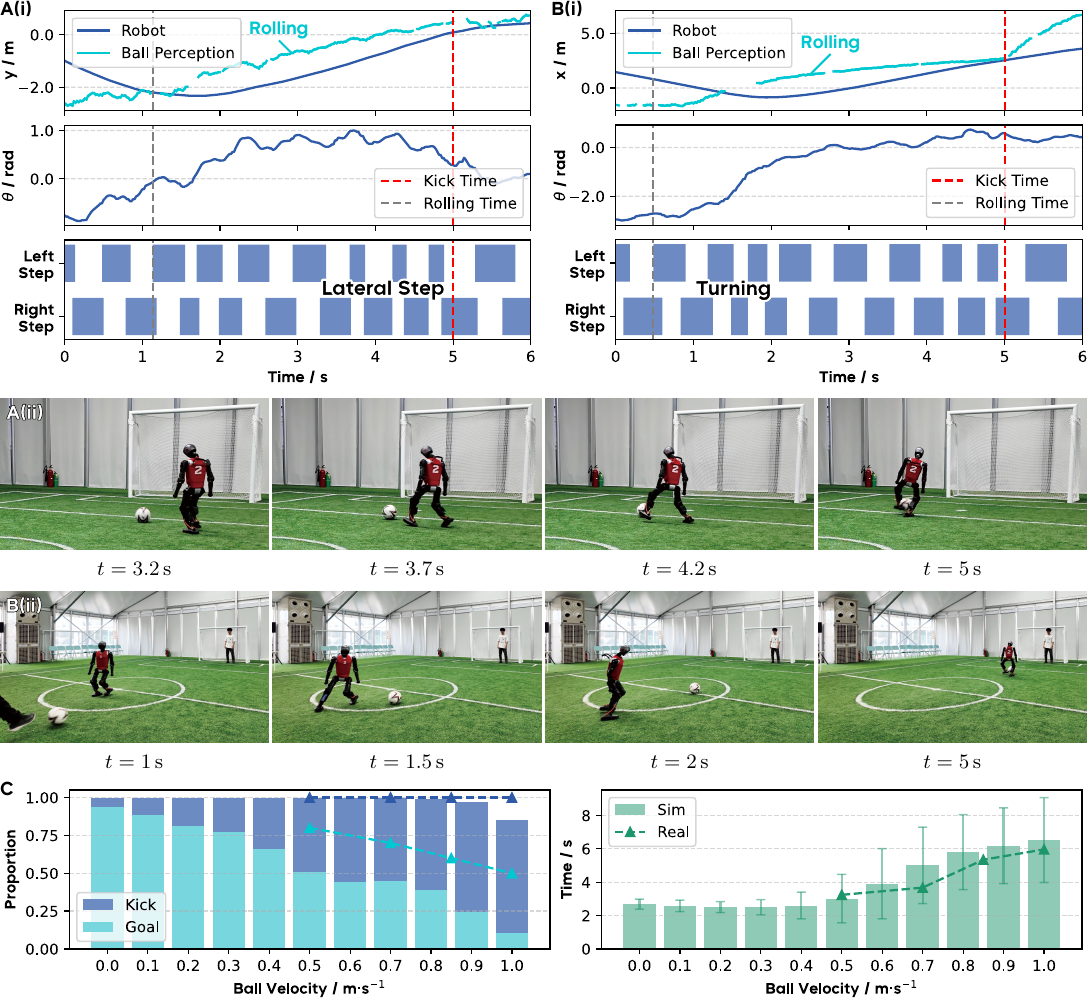}
    \caption{\textbf{Reactive kicking with rolling balls.} (\textbf{A}) Agile lateral movement and kicking when the ball rolls horizontally: (i) robot and ball trajectories together with the corresponding foot contact patterns and (ii) snapshots of the robot's behavior. (\textbf{B}) Agile turning and kicking when the ball rolls toward the robot's rear, with (i) and (ii) showing the same types of results as in (A). (\textbf{C}) Kick and goal-scoring success rates and time-to-kick versus ball velocity. Simulation results are shown as bars based on 8192 trials per velocity, and real-world results are shown as triangle markers based on 10 trials per velocity. Error bars indicate the SD across the simulation trials.}
    \label{fig:gait}
\end{figure}

This gait adjustment became even more pronounced when chasing a rolling ball. When the ball rolled in front of the robot, the policy triggered rapid lateral steps in an attempt to intercept its trajectory before it moved out of range (Fig.~\ref{fig:gait}A). Furthermore, agile turning behavior was exhibited when the ball rolled from the robot's side toward its rear (Fig.~\ref{fig:gait}B). In such scenarios, the robot executed a rapid rotational maneuver, where one foot pushed off the ground while the other pivoted, enabling full reorientation in as few as two steps, thereby retaining visual contact with the ball. To quantitatively evaluate the policy's responsiveness and robustness to rolling balls, we conducted real-world experiments in which the robot started at the field center while the ball rolled laterally from a position 3~m in front of and 1.5~m to the right of the robot at different initial speeds. Because of unavoidable friction in real-world settings, the ball velocity inevitably decayed during pursuit. To obtain a more systematic characterization, we further performed simulations with constant ball speeds ranging from 0.1 to 1.0~m/s under otherwise identical conditions. For direct comparison, the real-world results were aligned with simulation according to the approximate average ball velocity during pursuit, as summarized in Fig.~\ref{fig:gait}C. The policy maintained stable performance for rolling speeds below 0.5~m/s, exhibiting no substantial increase in time-to-kick and retaining a success rate of more than 50\%. At higher speeds, the increased difficulty of rapid pursuit and precise foot placement led to longer interception times and a gradual decline in success rate. Real-world experiments followed trends consistent with simulation and exhibited comparable response times, indicating effective sim-to-real transfer. The higher success rates observed on hardware were attributed to velocity decay during pursuit, which reduced the difficulty of interception. Collectively, these results demonstrated the policy's ability to reactively couple visual feedback of ball motion with agile locomotion and precise kicking under dynamic conditions.

\section{Discussion}

In this work, we enabled reactive soccer skills for humanoid robots in challenging real-world environments through an RL-based controller that tightly integrated visual perception with agile locomotion. Building on the AMP framework, our approach incorporated perceptual feedback into the learning process, allowing the policy to acquire versatile soccer behaviors from human demonstrations while responding reactively to visual inputs. To achieve this, we introduced an encoder-decoder architecture together with a virtual perception system based on a compact perception abstraction that promoted consistent visual representations across simulation and the real world. These components effectively bridged the mismatch between noisy real-world perception and the precise motion control required for stable kicking, enabling robust motion imitation under real-world constraints and reliable policy execution across diverse environments.

The practical efficacy of the proposed controller was confirmed through experiments across a range of real-world conditions. The humanoid robot maintained stable and fluent soccer performance across varied terrains, adapting seamlessly to variations in interaction dynamics and visual noise while exhibiting high agility and responsiveness. This adaptability manifested in dynamic movement adjustments to varying ball positions and trajectories, with the robot achieving faster ball possession compared with rule-based strategies and reactively responding to rolling balls. This superiority was further validated in the RoboCup Humanoid League competition, where the robot successfully scored goals under strict constraints. These results underscore the practical value of our approach in real-world competitive settings.

The use of an adversarial discriminator provides implicit motion guidance for the policy to acquire soccer skills aligned with human motion patterns while reducing reliance on extensive reward engineering. The resulting unified policy exhibits versatile behaviors while preserving the flexibility required for dynamic scenarios. This unified policy not only simplifies the system architecture but also promotes seamless coordination between perception and locomotion, as evidenced by the robot's ability to dynamically adjust its gait, head orientation, and kicking angle in response to real-time ball and goal positions. In contrast with conventional rule-based pipelines that separate strategic planning from motor control, our approach avoids the disjointed movements produced by modular systems. It also circumvents the rigidity of feature-based imitation methods that rely on strict temporal alignment with reference motions and thus lack adaptability in dynamic scenarios.

Robust performance across environments is further supported by the deliberate design of the observation space and perception pipeline. Instead of conditioning the policy on high-dimensional raw images, we used a compact state space derived from ball detections, providing an abstract and consistent representation across simulation and hardware, thereby enhancing generalization to diverse real-world conditions. An odometry module additionally provides long-term memory of the robot's location, improving robustness against temporary losses of visual landmarks. During training, the integration of the virtual perception system with the encoder-decoder architecture encourages the policy to actively coordinate perception and action. This interaction allows the controller to mitigate perception noise and maintain accurate ball position estimates during motion, improving robustness to motion blur, partial observability, and other perceptual disturbances in dynamic soccer scenarios. The emergence of purposeful ball-search and active tracking behaviors, together with the reduction in position estimation errors, further demonstrates how the perception pipeline enables the policy trained in simulation to operate reliably in real-world environments.

Despite these successes, our work has several limitations that warrant further exploration. First, the current controller focuses primarily on individual soccer skills and lacks the capability to coordinate with teammates or respond to opponents in multirobot scenarios. The policy's observations are currently limited to the ball and goal, without incorporating information about other agents in the environment. This restricts its ability to perform complex team behaviors such as collaborative passing, defensive interception, or dynamic role allocation. A natural direction for future work is to extend the training framework to multiagent environments and augment the observation space with real-time observations of other robots, enabling dynamic cooperation and adversarial adaptation in complex team-based soccer scenarios. Second, although the policy demonstrates reliable walking and adaptive kicking across varied ball and goal configurations, it still falls short of reproducing the broader range of soccer behaviors observed in human play, such as dribbling, ball trapping, or strategic passing. This gap stems primarily from the policy's limited ability to identify context-specific action requirements across different situations, such as determining when to prioritize maintaining possession of the ball rather than striking it. The current training objective, which emphasizes goal-scoring outcomes, tends to drive the policy toward a limited set of movement patterns. Incorporating strategic priors or task-level decision cues, analogous to the tactical planning used by human players, could promote the emergence of richer and more context-aware behaviors, bringing the robot's performance closer to human-level capabilities.

In conclusion, this study demonstrates that incorporating perceptual characteristics into the learning framework enables humanoid robots to acquire vision-driven soccer skills that transfer robustly from simulation to the real world. As humanoid robots move toward deployment in dynamic and unstructured environments, the ability to integrate perception with motion control will become increasingly essential for effective interaction with the physical world. Beyond the specific domain of soccer, our results suggest an approach for advancing real-world humanoid capabilities through the integrated learning of perception and action, bringing robots closer to autonomous and adaptive embodied systems.

\section{Materials and Methods}

\begin{figure}[!t]
    \centering
    \includegraphics[width=\textwidth]{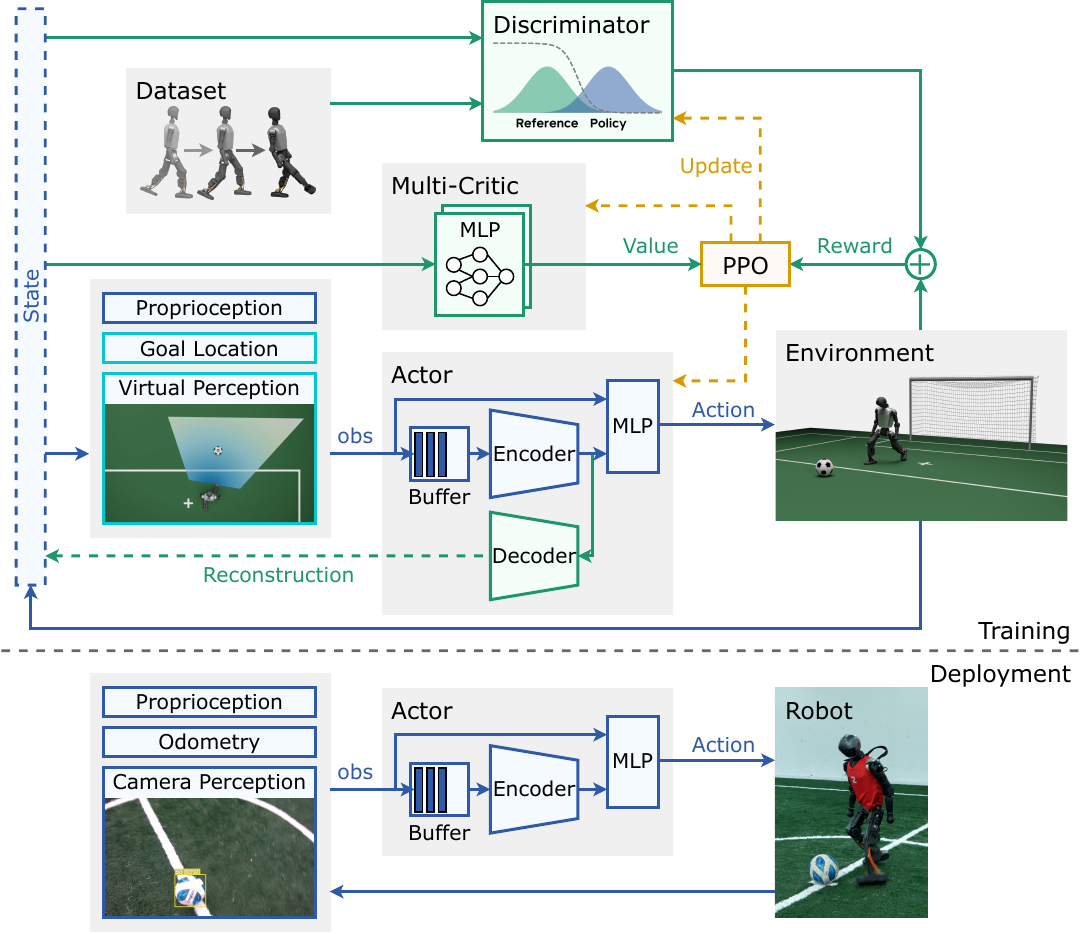}
    \caption{\textbf{The proposed learning framework.} The actor receives partial observations and reconstructs the full state from historical data using an encoder-decoder architecture. The policy is trained with PPO, using rewards from both the environment and a discriminator encoding motion priors, with multiple critics providing value estimates. The policy is trained in simulation with only the modules highlighted in blue deployed on hardware.}
    \label{fig:framework}
\end{figure}

Our main objective was to develop a system that enabled humanoid robots to kick the ball into the goal with onboard vision. An overview of our method is presented in Fig.~\ref{fig:framework}.

\subsection{Training environment}

We trained the policy using RL in simulation and then transferred it to the real world. The simulation environment, constructed using NVIDIA Isaac Gym, comprised a robot and a ball on a soccer field. The field dimensions adhered to the adult-size specifications of the RoboCup Humanoid League, measuring 14~m in length and 9~m in width, with two goals (each 2.6~m in width) positioned at opposite ends. With reference to the general characteristics of standard size 5 soccer balls, we randomized the ball mass, friction, and restitution during training to capture diverse ball-robot and ball-terrain interaction dynamics, with detailed parameter ranges provided in the Appendix. Despite the flat terrain of the deployed real-world field, mild terrain unevenness was incorporated during the training process to facilitate more stable locomotion in real-world scenarios while enriching the diversity of ball-terrain contact dynamics.

At the start of each episode, the robot and the ball were randomly initialized in the field. An episode terminated when the robot fell, the ball went out of bounds, a goal was scored, or a time limit of 60~s was reached. To facilitate the robot's learning of continuous kicking skills, when the ball went out of bounds or entered the goal, only the ball's position was reset, whereas the robot's state remained unchanged. To simulate potential scenarios in real matches, the ball was randomly subjected to an additional velocity or teleported to another position on the field with a certain probability, thereby mimicking interventions on the ball by opponents or referees. Similarly, external forces or extra velocities were applied to the robot to simulate physical confrontations during gameplay. These configurations contributed to the development of robust behaviors.

\subsection{Training procedure}

We formulated the control problem as a partially observable Markov decision process (POMDP) and adopted asymmetric actor-critic \cite{pinto2017asymmetric} and proximal policy optimization (PPO) \cite{schulman2017proximal} to train the policy. The actor and critic were represented by a policy network and a value network, parameterized by the multilayer perceptron (MLP). The value network was trained to estimate the state value, receiving the full state available only in the simulator. The policy had access only to partial information about the complete state, which was consistent with information that could be obtained from sensors on the robot in the real world, and output desired joint positions as actions that were tracked using a proportional-derivative controller.

Although the policy's observations were noisy and incomplete, critical latent information could be inferred from historical observations \cite{lee2020learning,li2023robust}. To leverage this temporal context, we provided the network with a time sequence of 50 preceding observation frames (covering 1~s of history) alongside the current observation. This sequence was encoded into a 64-dimensional latent representation using an MLP, which was then concatenated with the current observation and fed into the actor network. Following the approach in \cite{gu2024advancing}, we further integrated a decoder that reconstructed auxiliary information critical to the task but unavailable during real-world deployment from this latent state, such as the robot's dynamic parameters and the ball's true position. This design guided the policy to extract meaningful insights from historical data, facilitating the learning of more robust latent representations. By implicitly modeling the mapping between noisy observations and their underlying true states, the framework enhanced the policy's ability to perform accurate state estimation and adaptive behavior in real-world deployment.

The agent's objective was to kick the ball into the goal. To facilitate the learning of long-horizon kicking behaviors, we addressed the high sparsity of goal-scoring rewards by introducing two dense auxiliary rewards in addition to the terminal goal reward. One encouraged the robot to approach the ball, and another promoted the ball's motion toward the goal. This design provided the agent with more frequent and informative feedback during training. The proximity rewards were constructed using potential-based reward shaping \cite{ng1999policy}, where the potential function was defined as the Euclidean distance between relevant entities, such as the robot and the ball or the ball and the goal. This formulation guaranteed that the learned behaviors remained directed toward successful goal scoring.

In our training setting, the goal-related reward was sparse and nonsmooth. When a single critic was used to estimate the combined return, the high variance and intermittency of the task reward introduced substantial noise into the shared value estimates, resulting in noisy and unreliable policy gradients and degraded learning performance, as observed in Fig.~\ref{fig:reconstruct_noise}A. In the later stages of training, the increasing magnitude and variability of the task reward further reduced the signal-to-noise ratio of policy updates and manifested as elevated policy entropy. To mitigate this issue, we adopted a multicritic framework \cite{mysore2022multi}, in which separate critics independently estimated returns for two reward groups. The first group comprised goal-related rewards, and the second included auxiliary rewards such as style and regularization terms. Each reward group was treated as a dedicated subtask paired with its own critic, and a weighted sum of their respective advantages was then used by PPO to update the policy. This design allowed each advantage signal to be accurately estimated and independently normalized, preserving balanced gradient contributions across reward terms regardless of disparities in scale and sparsity \cite{zargarbashi2024robotkeyframing}, thereby mitigating interference between reward components.

\subsection{Perception system}

Incorporating perception into RL training amplified the challenges of sim-to-real transfer. Raw RGB (red, green, and blue) camera inputs introduced a substantial gap between simulation and the real world because of variations in lighting, texture, and noise characteristics. Although depth images \cite{zhuang2023robot,zhuang2024humanoid} and elevation maps \cite{hoeller2024anymal,wang2025beamdojo} proved effective for terrain traversal by capturing geometric abstractions, they were less suited to dynamic soccer scenarios, where accurate tracking of moving objects is essential. Alternative approaches that learned from synthetic RGB images \cite{tirumala2024learning,yu2024learning} in simulation demanded extensive computational resources to cover the diverse visual conditions of real environments, which limited scalability. To overcome these challenges, following prior work \cite{ji2023dribblebot}, we adopted an object-centric perception abstraction that extracted task-specific structured information from visual inputs, which proved effective in narrowing the sim-to-real gap while preserving task relevance. After perceptual processing, the policy received only the ball position with a ball detection flag, the goal center coordinates, and the goal normal direction, all expressed in the robot's coordinate frame. The detection flag was a binary indicator of ball visibility, with the ball position set to zero when the ball was not detected. This representation enabled the execution of vision-driven behaviors and abstracted away irrelevant visual complexity. By focusing solely on environmental cues critical to soccer performance, our framework promoted the development of robust, generalizable behaviors with substantially reduced training overhead.

Specifically, we used the You Only Look Once object detection model \cite{yolov8_ultralytics}, fine-tuned on a self-collected dataset from the robot's onboard camera, to detect the ball and field landmarks such as goalposts and line intersections from RGB images. Further details are provided in the Appendix. The pixel coordinates of these detected features in 2D images were projected into the BEV space through a fusion of depth-based and geometric projections using the estimated camera height and pose. The BEV positions of field landmarks were then fused with proprioceptive odometry to achieve stable long-horizon localization, from which real-time estimates of the goal center and normal direction were derived. For ball position estimation, conventional approaches typically filtered detected ball positions and fused them with odometry data to mitigate perceptual noise and detection failures. However, these methods often required slow calibration when the robot approached the ball to achieve accurate estimates and struggled to cope with highly dynamic kicking motions. Instead, we enabled the policy to directly process noisy or incomplete perceptual observations, allowing it to autonomously learn the optimal balance between perception and action for accurate ball positioning during kicking.

Although ground-truth ball positions were available in simulation, we introduced a virtual perception system in the simulation environment to expose the policy to realistic perceptual characteristics during training. This system replicated camera and perception processing behaviors observed on the real robot by modeling four key perceptual factors: detection probability, noise distribution, update frequency, and latency. Compared with prior work \cite{ji2023dribblebot}, we explicitly modeled the dependence of perceptual noise on ball distance and incorporated FOV constraints relative to head pose, enabling the policy to learn active perception through head motion and to recover reliable, low-noise information from imperfect visual data via the encoder-decoder architecture. These extensions addressed key challenges in agile motion, partial observability, and long-horizon localization that were particularly pronounced for humanoid soccer.

To construct the virtual perception model, we collected data from the real robot observing the ball under diverse conditions, including varying head orientations and relative positions. A rule-based ball-tracking program, executed using the robot's default walking gait, was deployed while a human operator moved the ball across the field to collect $\sim1$ hour of data. Ground-truth positions of both the robot and the ball were obtained via a motion capture system, alongside logs of visually detected ball positions and the robot's joint positions. Positional noise was modeled using a Gaussian distribution, with variance regressed as a linear function of the relative position between the robot and the ball. In addition, we estimated the robot's FOV and corresponding detection probabilities, as well as the perception update frequency and latency between image capture and detection output. These modeling efforts were essential for achieving accurate and reliable kicking behaviors on the physical robot. Notably, although the walking gait used for data collection differed from the learned policy, which introduced potential distribution discrepancies, the simplicity of our perceptual modeling ensured strong generalization across locomotion patterns.

\subsection{Learning from demonstration}

To improve the robot's kicking performance, we drew inspiration from human kicking behaviors, where the arch of the foot was the most frequently used part. The arch provided a larger contact area with the ball compared with other parts of the foot, such as the curved toe, facilitating better ball control and higher kicking accuracy. This characteristic avoided large deviations in kicking direction caused by minor positional errors when using less-stable contact points. To encourage the policy to learn humanlike kicking behaviors and this specific kicking pattern, the AMP approach \cite{peng2021amp} was used to specify the agent's behavioral style by leveraging unstructured motion clip datasets. This method enabled automatic balancing between task execution and motion imitation and could dynamically interpolate and generalize from the dataset without explicit clip selection, ensuring that the robot performed the task effectively while exhibiting stylistically consistent movements.

Specifically, we first curated a dataset of reference motions encompassing relevant behaviors. We selected 76.28~s of omnidirectional walking data, covering forward, backward, turning, and side-stepping motions, from the Advanced Computing Center for the Arts and Design motion capture dataset \cite{ACCAD} and recorded 30~s of arch-based kicking motions using an optical motion capture system. These human movements were retargeted to the robot to align with its physical dimensions and joint configuration using an optimization-based retargeting method detailed in the Appendix. This dataset was used to train a discriminator network, which was adversarially optimized to distinguish state transitions $(\boldsymbol{s}_t, \boldsymbol{s}_{t+1})$ from the reference motion dataset (real samples) and those generated by the robot during training (fake samples), thereby learning a general metric for similarity between the robot's motions and the stylistic characteristics embedded in the dataset. The discriminator's predictions were used to compute a style reward that guided the robot toward movement patterns consistent with the reference dataset. The policy was trained via RL to maximize a cumulative reward that combined this style reward with other environmental rewards. This framework enabled the policy to autonomously compose skills, such as transitioning from walking to leg positioning for kicking, without the need for explicit motion sequencing.

To stabilize adversarial training dynamics and prevent mode collapse, we adopted a Wasserstein GAN with gradient penalty \cite{gulrajani2017improved} and applied a soft boundary constraint to the discriminator \cite{tang2024humanmimic}. In addition, we initialized the robot's states by randomly sampling from reference motion clips \cite{peng2018deepmimic}, which aided in stabilizing training during the initial phase. We further introduced a mirror symmetry loss \cite{yu2018learning} to promote symmetric motions, which was critical for facilitating the policy's acquisition of bilateral kicking capabilities and preventing convergence to a unilateral kicking strategy.

To enable reliable deployment on real humanoid hardware, we integrated AMP with our perception pipeline to handle partial observability and adopted accurate physical modeling together with domain randomization to account for sensing and actuation imperfections in the real world. These designs collectively extended AMP toward practical real-world control.

\subsection{Statistical analysis}

Statistical analyses were performed in Python using the NumPy library. Each simulation rollout or real-world trial was treated as an independent sample. Means and SDs were computed from independent samples where applicable. Error bars and shaded regions represent SD unless otherwise specified. Sample sizes and analysis details were specified in the corresponding figure captions.

\section*{Acknowledgments}

We thank Booster Robotics for providing the robot platform, experimental facilities, and technical support. We thank Beijing Beiao Group Corp. Ltd. for providing access to the testing venue. We thank Tinglong Zheng and Yunkang Cheng for assistance in data collection and experiments. This work was partly supported by STI 2030-Major Projects (nos. 2021ZD0201402 and 2021ZD0201401), the Beijing Natural Science Foundation (no. L243004), and the Tsinghua University Initiative Scientific Research Program (Student Academic Research Advancement Program: Zhuiguang Special Project, no. 20257020011).

\clearpage

\bibliographystyle{unsrtnat}
\bibliography{main}

\clearpage

\beginappendix

\section{Training details}

Our training pipeline was built upon a sim-to-real framework \cite{wang2025booster}, utilizing NVIDIA Isaac Gym for large-scale parallel simulation. To accelerate training, we employed a distributed setup with eight NVIDIA V100 GPUs. The policy was trained for 20,000 epochs across 16,384 parallel environments, taking approximately one day to complete. The training hyperparameters used in training were summarized in Table~\ref{tab:hyperparameter}. The domain randomization ranges were detailed in Table~\ref{tab:domain_randomization}. Observation spaces for the actor, critic, and discriminator, along with the state space used for reconstruction, were detailed in Table~\ref{tab:observation}. To match real-world sensing conditions, only the actor's observations, which were restricted to available information on the physical robot, were corrupted with simulated sensor noise.

\begin{table}[!h]
    \centering
    \caption{\textbf{Training hyperparameters}}
    \label{tab:hyperparameter}
    \begin{tabular}{l c}
        \toprule
        \textbf{Parameter} & \textbf{Value} \\
        \midrule
        Critic layer sizes & $(256,256,128)$ \\
        Actor layer sizes & $(256,256,128)$ \\
        Encoder layer sizes & $(1024,128)$ \\
        Decoder layer sizes & $(128,128)$ \\
        Network activation & ELU \\
        Number of learning epochs & 5 \\
        Minibatch size & 4 \\
        Learning rate & Adaptive \\
        Discount factor & 0.995 \\
        GAE lambda & 0.95 \\
        Desired KL & 0.01 \\
        Optimizer & Adam \\
        Entropy coefficient & 0.01 \\
        Reconstruction coefficient & 1 \\
        Discriminator coefficient & 1 \\
        Gradient penalty coefficient & 50 \\
        Symmetry coefficient & 10 \\
        \bottomrule
    \end{tabular}
\end{table}

\begin{table}
    \centering
    \caption{\textbf{Domain Randomization}}
    \label{tab:domain_randomization}
    \begin{tabular}{lll}
        \toprule
        \textbf{Parameter} & \textbf{Unit} & \textbf{Range} \\
        \midrule
        Action delay & ms & $\mathcal{U}(0, 20)$ \\
        Motor bias & rad & $\mathcal{U}(-0.05, 0.05)+\text{default}$ \\
        Motor stiffness & Nm/rad & $\mathcal{U}(0.95, 1.05)\times\text{default}$ \\
        Motor damping & Nms/rad & $\mathcal{U}(0.95, 1.05)\times\text{default}$ \\
        \midrule
        Torso mass & kg & $\mathcal{U}(0.95, 1.05)\times\text{default}$ \\
        Torso CoM & m & $\mathcal{U}(-0.05, 0.05)+\text{default}$ \\
        Link mass & kg & $\mathcal{U}(0.98, 1.02)\times\text{default}$ \\
        Link CoM & m & $\mathcal{U}(-0.005, 0.005)+\text{default}$ \\
        Foot friction & -- & $\mathcal{U}(0.0, 1.0)$ \\
        Foot compliance & -- & $\mathcal{U}(0.5, 1.5)$ \\
        Foot restitution & -- & $\mathcal{U}(0.0, 1.0)$ \\
        \midrule
        Ball mass & kg & $\mathcal{U}(0.4, 0.45)$ \\
        Ball friction & -- & $\mathcal{U}(0.1, 0.3)$ \\
        Ball restitution & -- & $\mathcal{U}(0.0, 1.0)$ \\
        \bottomrule
    \end{tabular}
\end{table}

\begin{table}
    \centering
    \caption{\textbf{Observation and state spaces}}
    \label{tab:observation}
    \begin{tabular}{llcccc}
        \toprule
        \textbf{Observation} & \textbf{Description} & \textbf{Actor} & \textbf{Critic} & \textbf{Recon.} & \textbf{Disc.} \\
        \midrule
        Projected gravity & Gravity vector in the robot frame & \checkmark & \checkmark &  & \checkmark \\
        Angular velocity & Angular velocity of the robot's base from IMU & \checkmark & \checkmark &  & \checkmark \\
        Joint position & Joint position offset from default configuration & \checkmark & \checkmark &  & \checkmark \\
        Joint velocity & Joint velocity & \checkmark & \checkmark &  & \checkmark \\
        Previous action & Previous policy action output & \checkmark & \checkmark \\
        Ball position & Ball position $(x,y)$ in the robot frame & \checkmark & \checkmark & \checkmark \\
        Ball detection flag & Binary indicator of ball visibility & \checkmark & \checkmark \\
        Goal position & Goal position $(x,y)$ in the robot frame & \checkmark & \checkmark \\
        Goal direction & Goal direction $(\cos\theta,\sin\theta)$ in the robot frame & \checkmark & \checkmark \\
        \midrule
        Linear velocity & Linear velocity of the base in the robot frame & & \checkmark & \checkmark & \checkmark \\
        Base height & Height of the robot's base above the ground & & \checkmark & \checkmark \\
        Mass randomization & Randomized mass and CoM of the base link & & \checkmark & \checkmark \\
        Ball velocity & Ball velocity $(x,y)$ in the world frame & & \checkmark & \checkmark \\
        Ball friction & Friction force $(x,y)$ acting on the ball & & \checkmark & \checkmark \\
        Foot position & Relative position of the feet in the torso frame & & & & \checkmark \\
        \bottomrule
    \end{tabular}
\end{table}

\section{Reward functions}

The reward function was a weighted sum of task rewards, style rewards, and regularization rewards that jointly guided policy learning. Reward components were given in Table~\ref{tab:reward}.

\begin{table}
    \centering
    \caption{\textbf{Reward components}}
    \label{tab:reward}
    \begin{tabular}{lll}
        \toprule
        \textbf{Reward} & \textbf{Description} & \textbf{Weight} \\
        \midrule
        Survival & Constant reward for each time step the robot remains operational & 3 \\
        Termination & Penalty when robot falls & $-1000$ \\
        Stagnation & Penalty when robot remains nearly motionless for 1~s & $-100$ \\
        Goal scoring & Reward when ball enters goal area & 15 \\
        Ball approach & Reward for reduction in distance from robot to ball & 50 \\
        Goal progress & Reward for reduction in distance from ball to goal & 500 \\
        Head pitch alignment & Penalty for difference between head pitch and ball elevation & $-0.5$ \\
        Head yaw alignment & Penalty for difference between head yaw and ball direction & $-0.5$ \\
        \midrule
        AMP style & Reward for motions similar to reference via adversarial discriminator & 0.3 \\
        Sideways kick & Reward for sideways foot movement when in contact with the ball & 20 \\
        Forward kick & Penalty for forward foot movement when in contact with the ball & $-20$ \\
        Foot proximity & Penalty when feet are too close to each other & $-5$ \\
        \midrule
        Head action rate & Penalty for abrupt changes in head joint actions & $-15$ \\
        Leg action rate & Penalty for abrupt changes in leg joint actions & $-1$ \\
        Joint position limit & Penalty for joint positions exceeding limits & $-100$ \\
        Base acceleration & Penalty for excessive root accelerations & $-0.001$ \\
        Collision & Penalty for collision on body parts except the feet & $-100$ \\
        \bottomrule
    \end{tabular}
\end{table}

To improve training stability under heterogeneous reward sources, we employed a multicritic structure, in which distinct reward groups were assigned to separate value functions. The goal-related critic received rewards associated with task progress, including goal-scoring, ball-approach, and goal-progress terms, whereas the auxiliary critic was trained on all remaining rewards.

Each critic learned its own value estimate from its respective reward subset. The final policy advantage combined the two advantages via a weighted sum:
\begin{equation}
A_{\text{total}} = w_{\text{goal}}A_{\text{goal}} + w_{\text{aux}}A_{\text{aux}},
\end{equation}
where $w_{\text{goal}}=2$ and $w_{\text{aux}}=1$. This decomposition reduced negative interference between reward components and improved the stability of PPO updates.

\section{AMP training}

The AMP implementation was based on a Wasserstein GAN with gradient penalty objective \cite{gulrajani2017improved}, with an additional soft boundary constraint on the discriminator \cite{tang2024humanmimic}. The discriminator $D_\phi$ provided a score that estimated the Wasserstein distance between expert and policy-generated motion transition distributions, and the policy $\pi_\theta$ was trained to minimize this distance as part of its optimization objective.

At each iteration, the discriminator processed pairs of consecutive motion states $\boldsymbol{x}_t = (\boldsymbol{s}_t, \boldsymbol{s}_{t+1})$, sampled either from expert demonstrations $\boldsymbol{x}_t^{E}$ or from policy rollouts $\boldsymbol{x}_t^{\pi}$, and output a scalar score $D_\phi(\boldsymbol{x}_t)$. The discriminator loss was defined as:
\begin{equation}
\mathcal{L}_D = -\mathbb{E}\left(\tanh(0.4 D_\phi(\boldsymbol{x}_t^{E}))\right) + \mathbb{E}\Big(\tanh(0.4 D_\phi(\boldsymbol{x}_t^{\pi}))\Big),
\end{equation}
where the $\tanh(\cdot)$ transformation bounded the discriminator output, preventing excessively large values from destabilizing adversarial optimization during early training \cite{tang2024humanmimic}.

To enforce Lipschitz continuity, a gradient penalty term \cite{gulrajani2017improved} was added:
\begin{equation}
\mathcal{L}_{\text{grad}} = \mathbb{E}\left(\left(\|\nabla D_\phi(\hat{\boldsymbol{x}}_t)\| - 1 \right)^2\right),
\end{equation}
where $\hat{\boldsymbol{x}}_t = \alpha \boldsymbol{x}_t^{E} + (1-\alpha) \boldsymbol{x}_t^{\pi}$ denoted a random interpolation between expert and policy samples, and $\alpha \sim \mathcal{U}(0,1)$.

The discriminator provided an adversarial reward
\begin{equation}
r_{\text{amp}} = -\tanh(0.4 D_\phi(\boldsymbol{x}_t^{\pi})),
\end{equation}
encouraging the policy to generate motions closely aligned with expert demonstrations.

\begin{figure}[!t]
    \centering
    \includegraphics[width=\textwidth]{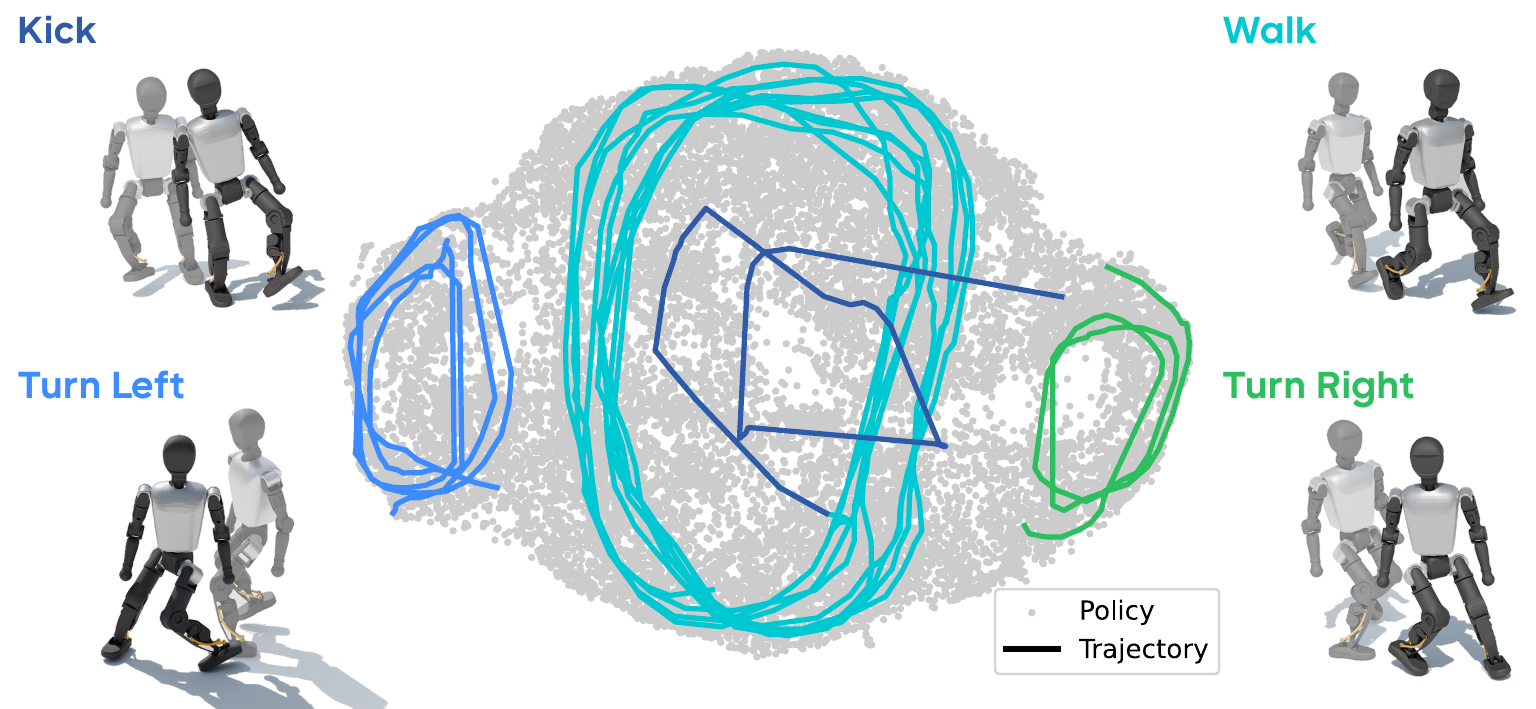}
    \caption{\textbf{UMAP visualization of policy behaviors without AMP.} Projection of 20,000 joint-space trajectory frames into a 2D latent space using UMAP. Representative trajectories corresponding to walking, turning, and kicking are highlighted to illustrate the distribution of different motion patterns.}
    \label{fig:umap_wo_amp}
\end{figure}

We conducted ablation studies to validate the effect of AMP on policy learning and real-world performance, and to isolate the contributions of its components. To examine the role of AMP in shaping the learned motion manifold, we visualized the joint-space trajectories produced by the policy trained without AMP using UMAP. In contrast to the well-separated clusters formed by the AMP-based policy in Fig.~\ref{fig:umap}, the embeddings obtained without AMP appeared considerably more compact and entangled, exhibiting no clear cluster separation, as shown in Fig.~\ref{fig:umap_wo_amp}. Highlighted representative trajectories further illustrated that although walking, turning, and kicking behaviors still emerged, the distinctions among these motion patterns became substantially blurred without AMP guidance. These observations suggested that AMP facilitated structured motion space and promoted well-organized behaviors.

In addition, we observed that the policy trained without AMP exhibited aggressive motions and rapid alternating foot contacts during kicking. Without sufficient motion constraints, the policy could converge to behaviors that achieved high task rewards in simulation while producing physically implausible motions. Such issues could in principle be mitigated through additional reward shaping and carefully designed regularization terms, but this typically required substantial task-specific engineering effort and iterative tuning. From this perspective, AMP served as an effective motion prior that encouraged physically plausible and structured behaviors while reducing reliance on extensive reward engineering \cite{escontrela2022adversarial}.

We further compared the least-squares GAN (LSGAN) adopted in the original AMP framework \cite{peng2021amp} with the Wasserstein GAN (WGAN) we employed. The policy trained with WGAN generated smoother and more humanlike motions. In contrast, although the LSGAN-based policy captured the overall motion style of the reference dataset, its resulting behaviors were noticeably less natural and exhibited reduced motion quality.

\begin{figure}[!t]
    \centering
    \includegraphics[width=0.6\textwidth]{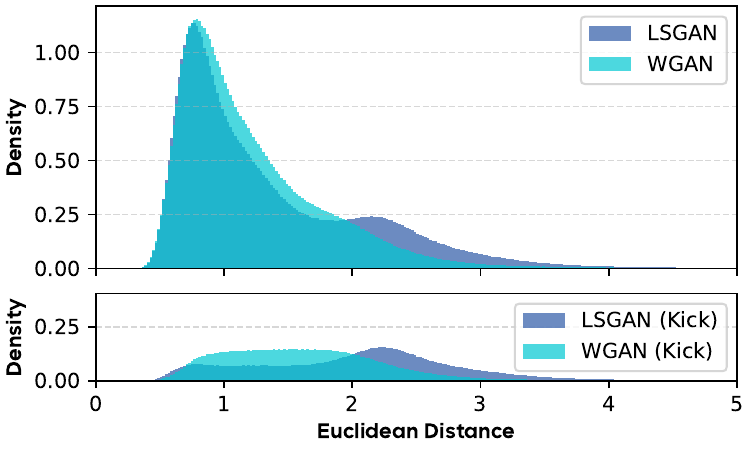}
    \caption{\textbf{Nearest-neighbor distance distributions for LSGAN and WGAN.} Distribution of Euclidean distances in the AMP observation space between policy-generated motions and the reference dataset. Motions are collected from 8192 parallel environments over 1000 steps for each policy.}
    \label{fig:amp_distance}
\end{figure}

To quantitatively evaluate the alignment between the learned motion and the reference dataset, we analyzed the distribution of nearest-neighbor distances in the AMP observation space. Specifically, for each frame generated by the policy, we computed its Euclidean distance to the closest frame in the reference motion dataset and aggregated these minimum distances across all samples. As illustrated in Fig.~\ref{fig:amp_distance}, the policy trained with WGAN yielded a distribution strongly concentrated at small distances, indicating closer alignment with the reference motions. In contrast, the LSGAN-based policy exhibited a secondary peak at larger distances. Further analysis restricted to kicking motions, defined as frames where the ball lay within 3~m of the robot, revealed that these large-distance samples primarily stemmed from collapsed kicking behaviors. These results confirmed that WGAN effectively alleviated mode collapse, enabling faithful and balanced learning of diverse motion patterns.

\section{Symmetry loss}

To discourage asymmetric strategies, such as consistently kicking with a single leg, we introduced a symmetry loss that promoted bilateral coordination. For each sampled observation $\boldsymbol{o}_t$, we constructed a mirrored counterpart $\tilde{\boldsymbol{o}}_t = \mathcal{M}_o(\boldsymbol{o}_t)$ using a mirroring operator $\mathcal{M}_o(\cdot)$ that swapped left-right joint states and inverted lateral components of ball and goal observations in the robot-centric frame.

The policy output two actions:
\begin{equation}
\boldsymbol{a}_t = \pi_\theta(\boldsymbol{o}_t), \quad
\tilde{\boldsymbol{a}}_t = \pi_\theta(\tilde{\boldsymbol{o}}_t).
\end{equation}
Applying a corresponding mirroring operator $\mathcal{M}_a(\cdot)$ to $\tilde{\boldsymbol{a}}_t$, we defined the symmetry loss as:
\begin{equation}
\mathcal{L}_{\text{sym}} = \mathbb{E}\left(\left\|\boldsymbol{a}_t - \mathcal{M}_a(\tilde{\boldsymbol{a}}_t) \right\|^2\right).
\end{equation}
This regularization encouraged consistent responses to symmetric states, maintaining bilateral motion capability and enabling the policy to autonomously select either foot depending on task context.

\begin{figure}[!t]
    \centering
    \includegraphics[width=0.8\textwidth]{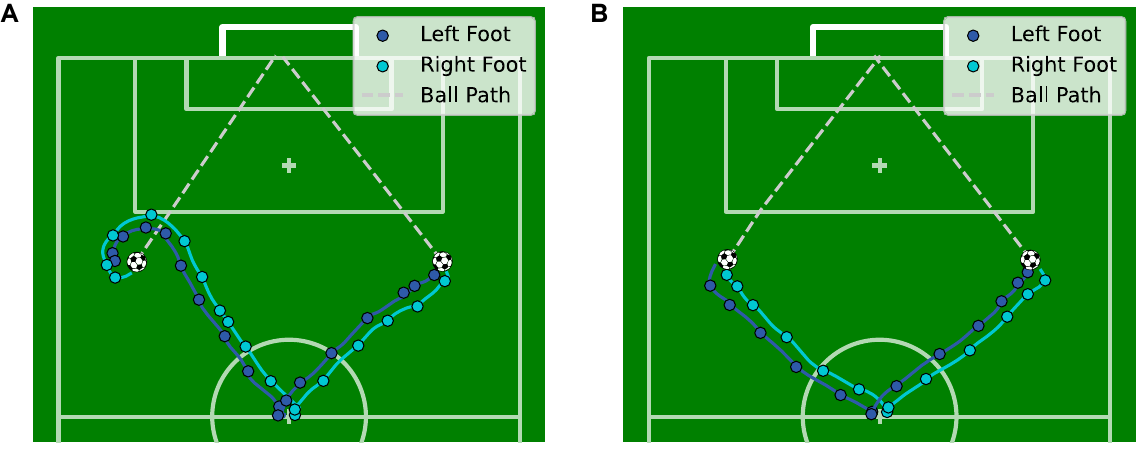}
    \caption{\textbf{Effect of mirror symmetry loss on kicking behavior.} For each policy, two trials were conducted with the ball placed on the left and right sides, respectively, with the robot starting from the field center. (\textbf{A}) Policy trained without symmetry loss. (\textbf{B}) Policy trained with symmetry loss.}
    \label{fig:mirror}
\end{figure}

We conducted an ablation study to evaluate the effect of the mirror symmetry loss. As shown in Fig.~\ref{fig:mirror}, incorporating this loss enabled the policy to adaptively select the kicking foot according to the ball position, yielding natural and efficient bilateral kicking behaviors. In contrast, removing the symmetry loss resulted in a strongly biased policy that predominantly relied on a single foot. When the ball lay on the opposite side, the robot first rotated around the ball to realign itself before kicking, leading to longer execution times and reduced overall efficiency.

\section{Perception system and modeling}

\begin{figure}[!t]
    \centering
    \includegraphics[width=0.6\textwidth]{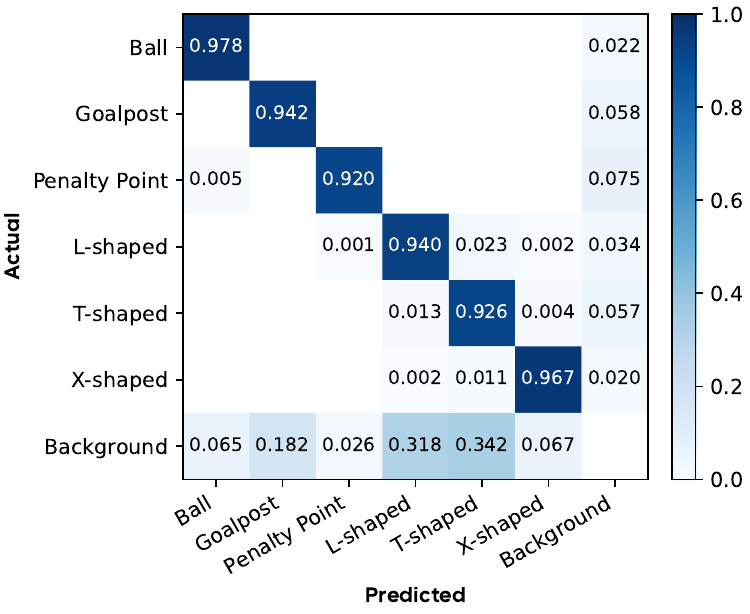}
    \caption{\textbf{Confusion matrix for the fine-tuned object detection model.} The model was evaluated on a test set containing 695 annotated image frames. Each row is normalized by the number of ground-truth instances in the corresponding category.}
    \label{fig:confusion_matrix}
\end{figure}

The object detection model was fine-tuned from YOLOv8s using a self-collected dataset containing approximately 30,000 annotated image frames captured from the robot's onboard camera at 3~Hz during robot operation. The detection categories included the ball, goalpost, penalty point, and T-, L-, and X-shaped field intersections. The confusion matrix of the fine-tuned model was shown in Fig.~\ref{fig:confusion_matrix}, demonstrating reliable detection performance across all categories. The remaining failure cases mainly occurred under severe motion blur or at extreme viewing distances, occasionally resulting in missed detections or confusion between visually similar field intersection types.

\begin{figure}[!t]
    \centering
    \includegraphics[width=\textwidth]{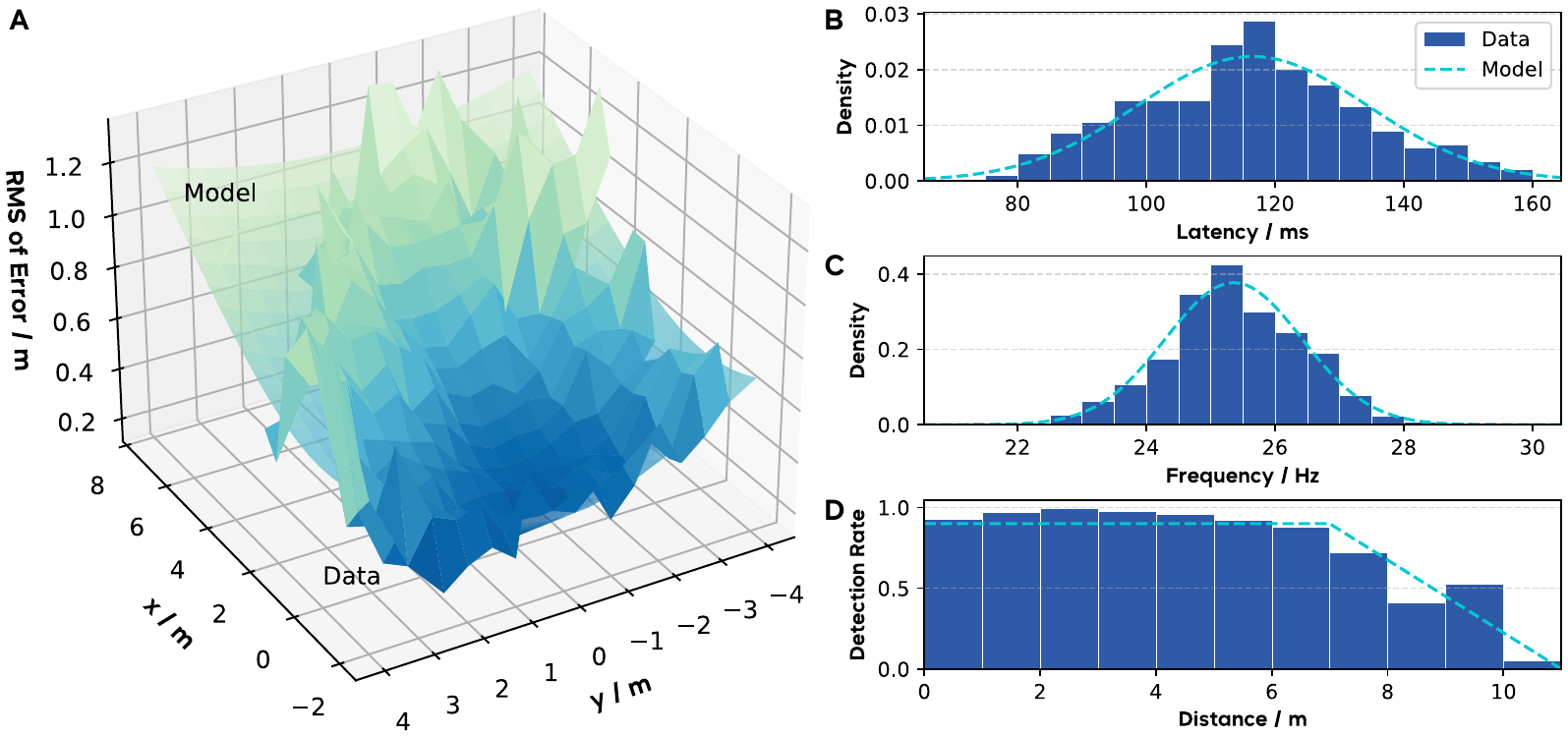}
    \caption{\textbf{Virtual perception system.} (\textbf{A}) Positional noise. (\textbf{B}) Distribution of perception latency. (\textbf{C}) Distribution of perception frequency. (\textbf{D}) Detection rate when the ball is in FOV.}
    \label{fig:noise_model}
\end{figure}

The modeling of our virtual perception system was illustrated in Fig.~\ref{fig:noise_model}. Positional noise was sampled from a Gaussian distribution $\mathcal{N}(0,(0.124\cdot d + 0.149)^2)$, where $d$ was the distance to the ball. Perception latency and update frequency were modeled as $\mathcal{N}(116~\text{ms},{(18~\text{ms})}^2)$ and $\mathcal{N}(25.36~\text{Hz},{(1.06~\text{Hz})}^2)$, respectively. The ball was detected with a 90\% probability within the robot's FOV up to a range of 7~m, beyond which the detection rate gradually decayed.

\begin{figure}[!t]
    \centering
    \includegraphics[width=0.6\textwidth]{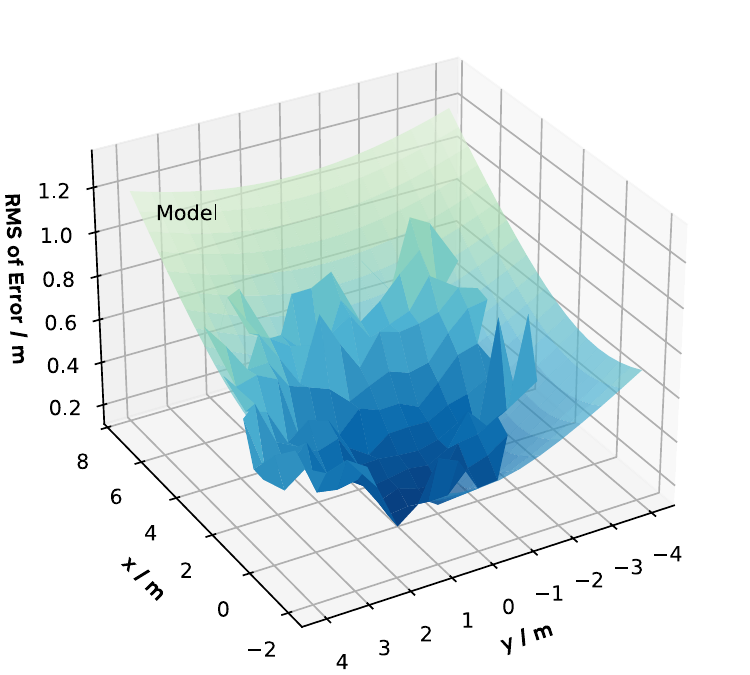}
    \caption{\textbf{Ball perception error distribution under policy execution.} Ball position errors measured over approximately 7000 perception frames are shown as a function of the ball location in the robot coordinate frame, together with the noise model used in the virtual perception system.}
    \label{fig:ball_noise}
\end{figure}

To validate the fidelity of the virtual perception model, we further evaluated the ball perception errors during real-world policy execution. We simultaneously recorded the ball positions perceived by the onboard camera and the corresponding ground-truth position obtained from a motion capture system, and analyzed the resulting error distribution, as shown in Fig.~\ref{fig:ball_noise}. The results indicated that the real-world perception error exhibited a distance-dependent trend that closely aligned with the noise characteristics assumed in the virtual perception system, supporting the validity of the proposed modeling. Notably, the overall magnitude of the measured errors was slightly lower than that of the model, which we attributed to the improved motion stability of the learned policy, leading to reduced camera jitter. These findings confirmed that the virtual perception system provided a faithful and robust approximation of real-world visual characteristics, enabling effective sim-to-real transfer.

\section{Odometry implementation}

We implemented a hybrid odometry system that fused proprioceptive and visual information to achieve robust and reliable localization.

The proprioceptive odometry module was trained using trajectories collected from policies executed in NVIDIA Isaac Gym. The simulator enabled large-scale data generation in parallel across diverse environmental conditions, allowing the model to learn generalizable mappings from proprioceptive inputs to the resulting motion. The odometry was implemented as an MLP that processed a 1~s temporal window of proprioceptive observations, including joint positions, joint velocities, base orientation, angular velocity, and the previous action. The previous odometry output was also fed back as part of the input, forming an autoregressive structure that enhanced robustness against accumulated drift. The network predicted the robot's incremental displacement over the next frame interval and was trained by minimizing the mean squared error between the predicted and ground-truth motion increments obtained from simulation.

To mitigate long-term drift, the proprioceptive estimates were corrected using vision-based localization derived from detected field landmarks, including goalposts, penalty points, and T-, L-, and X-shaped intersections. Each detected landmark was represented as a structured tuple comprising a categorical type label, a confidence score, and its 2D position expressed in the robot-centric BEV coordinate frame. A particle filter was employed to estimate the robot's global pose by matching these local observations against a predefined global field map. Each particle represented a pose hypothesis and was initialized within a constrained region centered around the prior estimate.

For each particle, detected landmarks were transformed from the local BEV frame to the global field frame via an SE(2) transformation, and their spatial consistency with the map was evaluated based on nearest-neighbor residuals. The particles were iteratively resampled with progressively reduced particle counts and noise variances, concentrating probability mass in low-residual regions of the pose space. The filter was deemed converged when the particle dispersion fell below predefined thresholds, and the particle with the minimum residual was selected as the final pose estimate. The estimate was then used to correct accumulated drift in the proprioceptive odometry, with the correction weighted by localization quality, reflected by the number of detected landmarks and the residual error.

Between visual updates, or when visual localization failed due to insufficient detected landmarks, excessive residuals, or lack of convergence within the maximum number of iterations, the system defaulted to pure proprioceptive odometry, ensuring continuous localization under temporary visual occlusions or degraded perception. This fusion strategy effectively compensated for proprioceptive drift while maintaining stable localization under intermittent loss of visual information.

\begin{figure}[!t]
    \centering
    \includegraphics[width=0.8\textwidth]{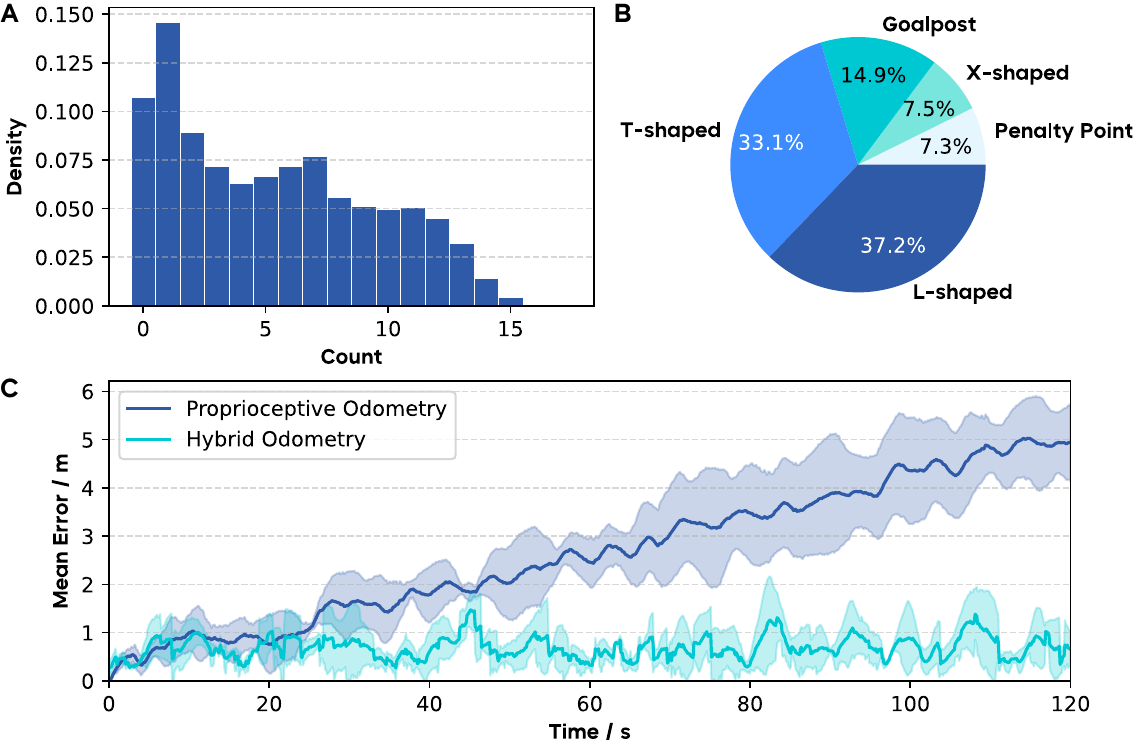}
    \caption{\textbf{Landmark detection and odometry evaluation.} (\textbf{A}) Distribution of the number of detected landmarks per frame, computed over 10,000 perception frames during real-world deployment. (\textbf{B}) Distribution of detected landmark categories over the same dataset. (\textbf{C}) Localization error of proprioceptive and hybrid odometry over 120~s of continuous policy execution. Shaded regions indicate the SD, computed from three trials.}
    \label{fig:landmark}
\end{figure}

To quantitatively assess the performance of the odometry module, we analyzed landmark detection statistics and localization accuracy during policy execution. The RoboCup field contained a total of 30 recognizable landmarks, comprising 12 T-shaped intersections, 10 L-shaped intersections, 4 goalposts, 2 X-shaped intersections, and 2 penalty points. We recorded landmark detections for each perception frame during real-world deployment. As shown in Fig.~\ref{fig:landmark}A, at least one landmark was successfully detected in approximately 89\% of all frames, and more than 5 landmarks were detected in about 52\% of frames. These frequent and diverse detections provided rich visual cues for reliable localization. Moreover, the distribution of detected landmark categories, illustrated in Fig.~\ref{fig:landmark}B, closely aligned with their actual proportions on the field, indicating balanced and stable detection performance across different landmark types.

To further evaluate localization accuracy, we recorded both the estimated robot trajectory and the ground-truth positions obtained from a motion capture system. As illustrated in Fig.~\ref{fig:landmark}C, purely proprioceptive odometry exhibited approximately linear error accumulation over time at a rate of about 0.04~m/s. In contrast, incorporating vision-based localization corrections effectively suppressed long-term drift, constraining the position error within approximately 1~m throughout policy execution and enabling stable, reliable localization.

\section{Deployment in RoboCup matches}

In RoboCup matches, robots had to operate under explicit game rules (e.g., kick-offs, free-kicks, and pauses) and coordinate with teammates, which introduced discrete game states and role-dependent behaviors. In practice, these components were commonly addressed through structured decision logic rather than continuous control policies.

To integrate the learned policy into this complex setting, we adopted a hierarchical control architecture in real matches. A high-level behavior tree governed game-level decision-making, including rule compliance, role assignment, and multi-robot coordination, and determined when to activate the learned policy based on the current game state and spatial relationships. When the learned policy was inactive, a velocity-tracking walking controller was used for navigation and positioning. This decomposition enabled the learned policy to be invoked in appropriate contexts throughout the match.

We further specified explicit triggering and termination conditions to ensure safe and effective activation of the learned policy. During gameplay, only the robot designated as the primary attacker, determined based on spatial advantage with respect to the ball, was permitted to enter the learned policy, and other robots executed supportive behaviors such as positioning, blocking, or passing using the walking controller. The attacker switched to the learned policy when the ball lay within a predefined distance threshold and the frontal space was sufficiently clear to avoid illegal collisions. The policy was terminated when the ball exited a predefined range or a maximum execution time was reached, after which the system transitioned back to the walking controller.

This hierarchical integration allowed the learned policy to exploit its strengths in tightly coupled perception-action control during critical ball-handling phases, with other components continuing to handle broader tactical and rule-based behaviors.

\section{Human motion retargeting}

To retarget human motions to the humanoid robot while respecting its distinct kinematic structure and physical dimensions, we adopted an optimization-based pipeline adapted from H2O \cite{he2024learning}, with modifications to accommodate our robot's morphology and support multiple human motion representations.

First, we established a semantic correspondence between the human body representation, in either BioVision Hierarchy (BVH) or Skinned Multi-Person Linear (SMPL) \cite{SMPL} format, and the robot's kinematic chain through a shared set of keypoints. Based on this correspondence, we fitted a set of body-mapping parameters to perform link-specific scaling and offset transformations on the human skeleton to align the limb lengths and body proportions between the human and the robot.

For each motion frame, we solved for the robot's joint angles, base position, and base orientation by minimizing a weighted sum of keypoint position errors, link orientation errors, and temporal smoothness regularization terms. The optimization was subject to joint limit constraints and ensured kinematic feasibility. Compared to the original H2O pipeline, we additionally optimized the base pose and introduced the link orientation and smoothness terms to improve motion quality. The resulting joint trajectories were further smoothed to ensure temporal continuity and execution feasibility.

This pipeline preserved the spatial structure of human movements while respecting the robot's physical constraints, yielding kinematically consistent reference motions for downstream policy training.

\end{document}